\documentclass{article}

     \PassOptionsToPackage{numbers, compress}{natbib}

     \usepackage[final]{neurips_2019}

\usepackage[utf8]{inputenc} 
\usepackage[T1]{fontenc}    
\usepackage{hyperref}       
\usepackage{url}            
\usepackage{booktabs}       
\usepackage{amsfonts}       
\usepackage{nicefrac}       
\usepackage{microtype}      

\usepackage{microtype}
\usepackage{graphicx}
\usepackage{subfig}
\usepackage[normalem]{ulem}

\usepackage{amsmath}

\allowdisplaybreaks[4]
\usepackage{amssymb}
\usepackage{graphics}
\usepackage{multirow}
\usepackage{float}
\usepackage{bm}
\usepackage{mathrsfs}
\usepackage{color,soul}
\usepackage{algorithm}
\usepackage{algorithmic}
\usepackage{enumerate}
\usepackage{todonotes}
\usepackage{latexsym}
\usepackage{wrapfig}
\usepackage{dsfont}
\sethlcolor{yellow}
\usepackage{diagbox}
\usepackage{stfloats}

\usepackage[export]{adjustbox}

\def\ddefloop#1{\ifx\ddefloop#1\else\ddef{#1}\expandafter\ddefloop\fi}

\newcommand\norm[1]{||#1||}

\def\ddef#1{\expandafter\def\csname v#1\endcsname{\ensuremath{\boldsymbol{#1}}}}
\ddefloop ABCDEFGHIJKLMNOPQRSTUVWXYZabcdefghijklmnopqrstuvwxyz\ddefloop

\def\ddef#1{\expandafter\def\csname v#1\endcsname{\ensuremath{\boldsymbol{\csname #1\endcsname}}}}
\ddefloop {alpha}{beta}{gamma}{delta}{epsilon}{varepsilon}{zeta}{eta}{theta}{vartheta}{iota}{kappa}{lambda}{mu}{nu}{xi}{pi}{varpi}{rho}{varrho}{sigma}{varsigma}{tau}{upsilon}{phi}{varphi}{chi}{psi}{omega}{Gamma}{Delta}{Theta}{Lambda}{Xi}{Pi}{Sigma}{varSigma}{Upsilon}{Phi}{Psi}{Omega}{ell}\ddefloop

\def\ddef#1{\expandafter\def\csname bb#1\endcsname{\ensuremath{\mathbb{#1}}}}
\ddefloop ABCDEFGHIJKLMNOPQRSTUVWXYZ\ddefloop

\title{Pareto Multi-Task Learning}

\author{
  Xi Lin\textsuperscript{1}, \ Hui-Ling Zhen\textsuperscript{1}, \ Zhenhua Li\textsuperscript{2}, \ Qingfu Zhang\textsuperscript{1}, \ Sam Kwong\textsuperscript{1} \\
  \textsuperscript{1}City University of Hong Kong, \ \textsuperscript{2}Nanjing University of Aeronautics and Astronautics \\
  \texttt{xi.lin@my.cityu.edu.hk, \ huilzhen@um.cityu.edu.hk, \ zhenhua.li@nuaa.edu.cn} \\
  \texttt{\{qingfu.zhang, cssamk\}@cityu.edu.hk}
}

\begin{document}

\maketitle

\begin{abstract}
Multi-task learning is a powerful method for solving multiple correlated tasks simultaneously. However, it is often impossible to find one single solution to optimize all the tasks, since different tasks might conflict with each other. Recently, a novel method is proposed to find one single Pareto optimal solution with good trade-off among different tasks by casting multi-task learning as multiobjective optimization. In this paper, we generalize this idea and propose a novel Pareto multi-task learning algorithm (Pareto MTL) to find a set of well-distributed Pareto solutions which can represent different trade-offs among different tasks. The proposed algorithm first formulates a multi-task learning problem as a multiobjective optimization problem, and then decomposes the multiobjective optimization problem into a set of constrained subproblems with different trade-off preferences. By solving these subproblems in parallel, Pareto MTL can find a set of well-representative Pareto optimal solutions with different trade-off among all tasks. Practitioners can easily select their preferred solution from these Pareto solutions, or use different trade-off solutions for different situations.  Experimental results confirm that the proposed algorithm can generate well-representative solutions and outperform some state-of-the-art algorithms on many multi-task learning applications.
\end{abstract}

\section{Introduction}

\begin{wrapfigure}{R}{0.41\linewidth}
	\begin{minipage}{\linewidth}
    \begin{figure}[H]
	    \centering
	    \includegraphics[width=1.0 \linewidth]{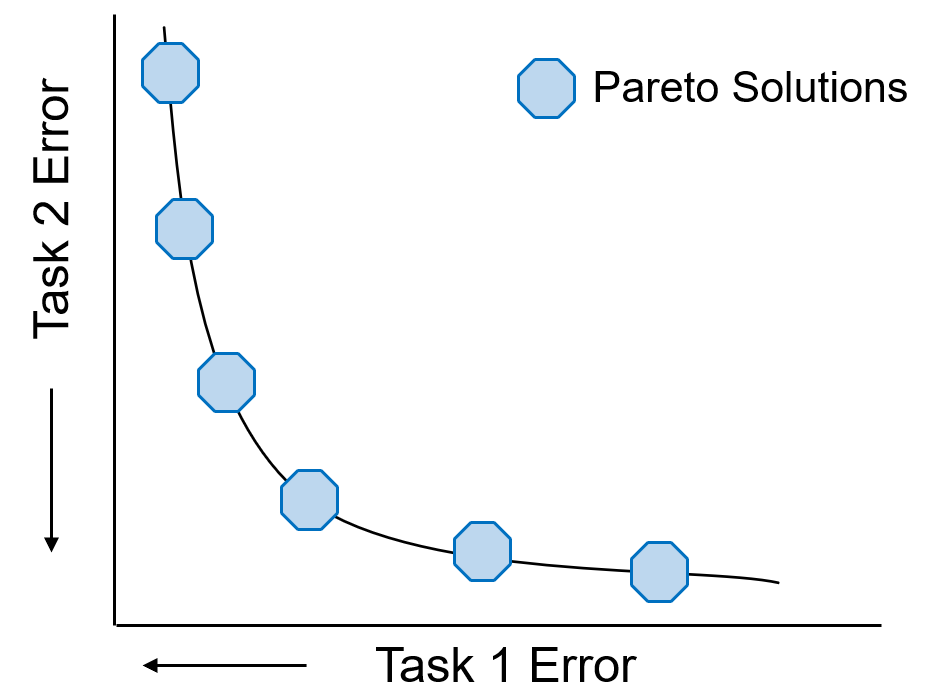}
	    \caption{\textbf{Pareto MTL} can find a set of widely distributed Pareto solutions with different trade-offs for a given MTL. Then the practitioners can easily select their preferred solution(s).}
        \label{MOPMTL}
    \end{figure}
	\end{minipage}
\end{wrapfigure}

Multi-task learning (MTL)~\cite{caruana1997multitask}, which aims at learning multiple correlated tasks at the same time, is a popular research topic in the machine learning community. By solving multiple related tasks together, MTL can further improve the performance of each task and reduce the inference time for conducting all the tasks in many real-world applications. Many MTL approaches have been proposed in the past, and they achieve great performances in many areas such as computer vision~\cite{kokkinos2017ubernet}, natural language processing~\cite{subramanian2018learning} and speech recognition~\cite{huang2015rapid}.

Most MTL approaches are proposed for finding one single solution to improve the overall performance of all tasks~\cite{ruder2017overview,zhang2017survey}. However, it is observed in many applications that some tasks could conflict with each other, and no single optimal solution can optimize the performance of all tasks at the same time~\cite{kendall2017multi}. In real-world applications, MTL practitioners have to make a trade-off among different tasks, such as in self-driving car~\cite{wang2018dels}, AI assistance~\cite{kim2017towards} and network architecture search~\cite{dong2018dpp,cai2018proxylessnas}.

\clearpage

How to combine different tasks together and make a proper trade-off among them is a difficult problem. In many MTL applications, especially those using deep multi-task neural networks, all tasks are first combined into a single surrogate task via linear weighted scalarization. A set of fixed weights, which reflects the practitioners' preference, is assigned to these tasks. Then the single surrogate task is optimized. Setting proper weights for different tasks is not easy and usually requires exhaustive weights search. In fact, no single solution can achieve the best performance on all tasks at the same time if some tasks conflict with each other.

Recently, Sener and Koltun~\cite{sener2018multi} formulate a multi-task learning problem as a multi-objective optimization problem in a novel way. They propose an efficient algorithm to find one Pareto optimal solution among different tasks for a MTL problem. However, the MTL problem can have many (even an infinite number of ) optimal trade-offs among its tasks, and the single solution obtained by this method might not always satisfy the MTL practitioners' needs.

In this paper, we generalize the multi-objective optimization idea~\cite{sener2018multi} and propose a novel Pareto Multi-Task Learning (Pareto MTL) algorithm to generate a set of well-representative Pareto solutions for a given MTL problem. As shown in Fig.~\ref{MOPMTL}, MTL practitioners can easily select their preferred solution(s) among the set of obtained Pareto optimal solutions with different trade-offs, rather than exhaustively searching for a set of proper weights for all tasks.

The main contributions of this paper are: \footnote{The code is available at: \url{https://github.com/Xi-L/ParetoMTL}}

\begin{itemize}
\item We propose a novel method to decompose a MTL problem into multiple subproblems with different preferences. By solving these subproblems in parallel, we can obtain a set of well-distributed Pareto optimal solutions with different trade-offs for the original MTL.
\item We show that the proposed Pareto MTL can be reformulated as a linear scalarization approach to solve MTL with dynamically adaptive weights. We also propose a scalable optimization algorithm to solve all constrained subproblems with different preferences.
\item Experimental results confirm that the proposed Pareto MTL algorithm can successfully find a set of well representative solutions for different MTL applications.
\end{itemize}

\section{Related Work}

Multi-task learning (MTL) algorithms aim at improving the performance of multiple related tasks by learning them at the same time. These algorithms often construct shared parameter representation to combine multiple tasks. They have been applied in many machine learning areas. However, most MTL algorithms mainly focus on constructing shared representation rather than making trade-offs among multiple tasks~\cite{ruder2017overview,zhang2017survey}.

Linear tasks scalarization, together with grid search or random search of the weight vectors, is the current default practice when a MTL practitioner wants to obtain a set of different trade-off solutions. This approach is straightforward but could be extremely inefficient. Some recent works~\cite{kendall2017multi,chen2018grad} show that a single run of an algorithm with well-designed weight adaption can outperform the random search approach with more than one hundred runs. These adaptive weight methods focus on balancing all tasks during the optimization process and are not suitable for finding different trade-off solutions.

Multi-objective optimization~\cite{miettinen2012nonlinear} aims at finding a set of Pareto solutions with different trade-offs rather than one single solution. It has been used in many machine learning applications such as reinforcement learning~\cite{van2014multi}, Bayesian optimization~\cite{zuluaga2013active,hernandez2016predictive,shah2016pareto} and neural architecture search~\cite{dong2018dpp,elsken2018efficient}. In these applications, the gradient information is usually not available. Population-based and gradient-free multi-objective evolutionary algorithms~\cite{zitzler1999evolutionary, Deb2001} are popular methods to find a set of well-distributed Pareto solutions in a single run. However, it can not be used for solving large scale and gradient-based MTL problems.

Multi-objective gradient descent~\cite{desideri2012mutiple,fliege2016method,fliege2000steepest} is an efficient approach for multi-objective optimization when gradient information is available.  Sener and Koltun~\cite{sener2018multi} proposed a novel method for solving MTL by treating it as multi-objective optimization. However, similar to the adaptive weight methods, this method tries to balance different tasks during the optimization process and does not have a systematic way to incorporate trade-off preference. In this paper, we generalize it for finding a set of well-representative Pareto solutions with different trade-offs among tasks for MTL problems.

\section{Multi-Task Learning as Multi-Objective Optimization}

\begin{figure*}[t]
\centering
\subfloat[Random Linear Scalarization]{\includegraphics[width = 0.33\textwidth]{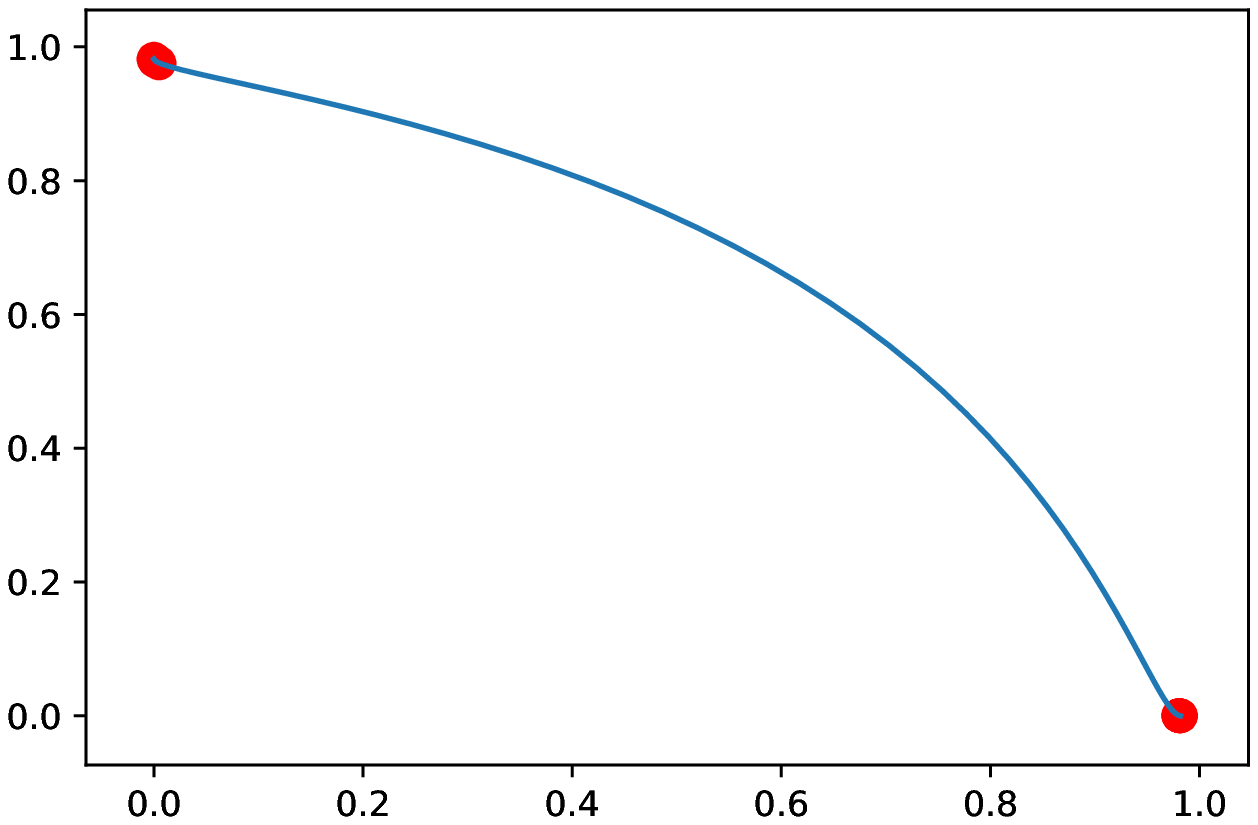}}
\subfloat[MOO MTL]{\includegraphics[width = 0.33\textwidth]{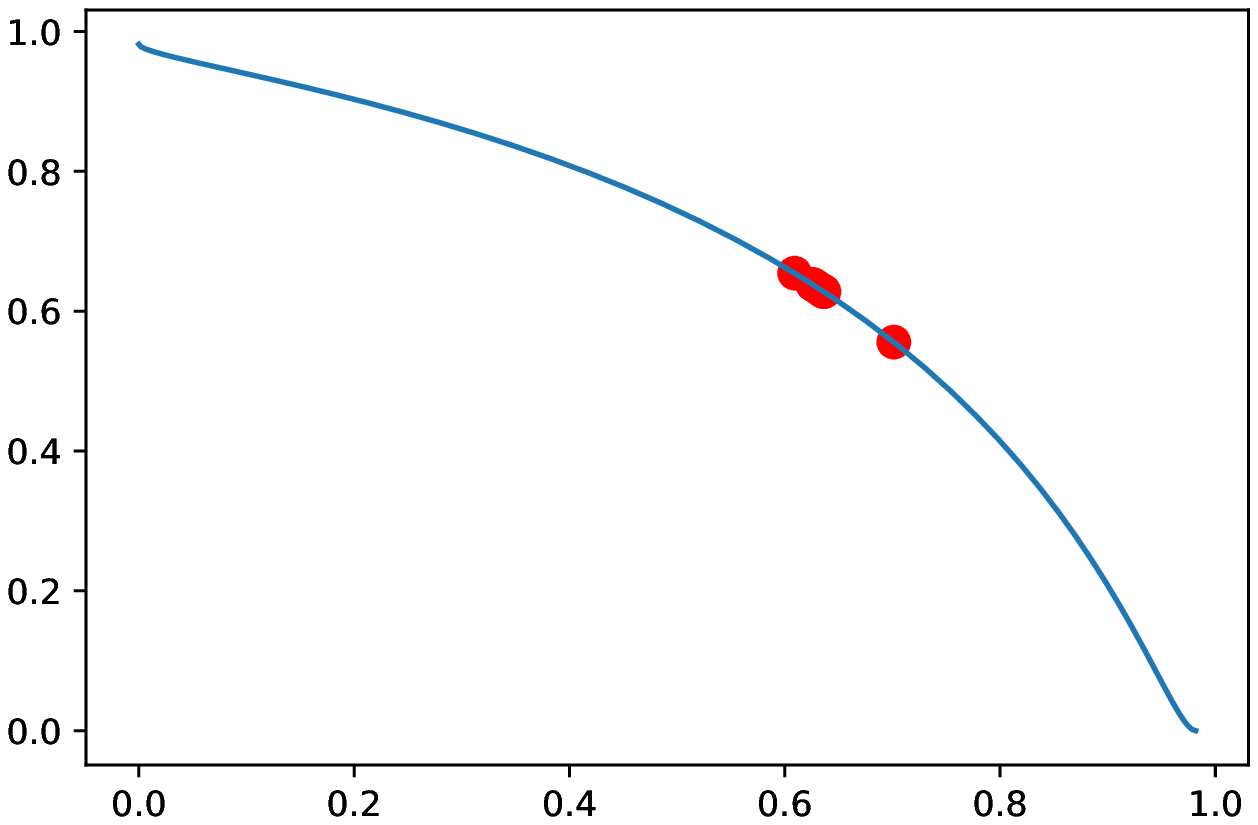}}
\subfloat[Pareto MTL (Ours)]{\includegraphics[width = 0.33\textwidth]{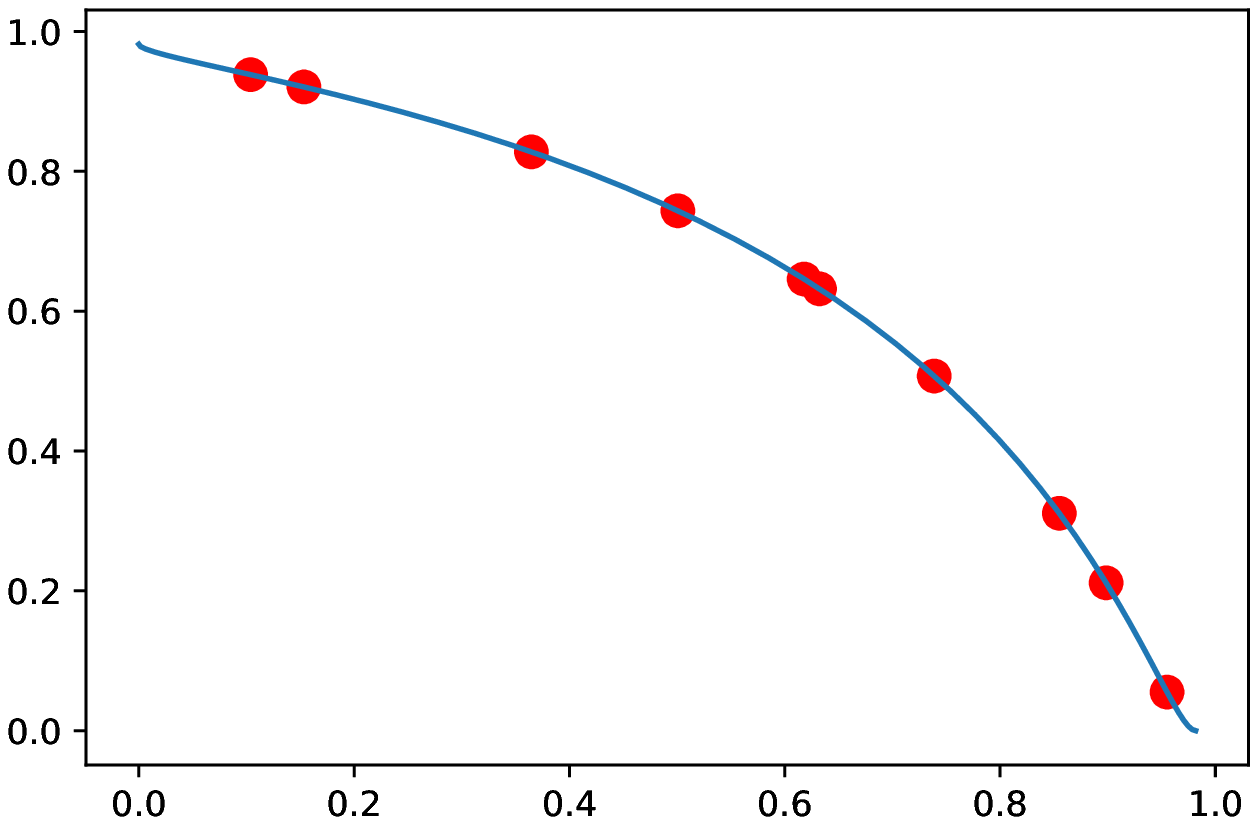}}
\caption{The convergence behaviors of different algorithms on a synthetic example. (a) The obtained solutions of random linear scalarization after 100 runs. (b) The obtained solutions of the MOO-MTL~\cite{sener2018multi} method after 10 runs. (c) The obtained solutions of the Pareto MTL method proposed by this paper after 10 runs. The proposed Pareto MTL successfully generates a set of widely distributed Pareto solutions with different trade-offs. Details of the synthetic example can be found in section 5.}
\label{PF_toy_example}
\end{figure*}

\subsection{MTL as Multi-Objective Optimization}

A MTL problem involves a set of $m$ correlated tasks with a loss vector:
\begin{align}\label{eq:mop}
\mathrm{min}_{\theta} \ \mathcal{L}(\theta) = (\mathcal{L}_1(\theta),\mathcal{L}_2(\theta),\cdots,
\mathcal{L}_m(\theta))^{\mathrm{T}},
\end{align}
where $\mathcal{L}_i(\theta)$ is the loss of the $i$-th task. A MTL algorithm is to optimize all tasks simultaneously by exploiting the shared structure and information among them.

Problem~(\ref{eq:mop}) is a multi-objective optimization problem. No single solution can optimize all objectives at the same time. What we can obtain instead is a set of so-called Pareto optimal solutions,  which provides different optimal trade-offs among all objectives. We have the following definitions~\cite{zitzler1999multiobjective}:

\textbf{Pareto dominance.} Let $\theta^a,\theta^b$ be two points in $\Omega$, $\theta^a$ is said to dominate $\theta^b$ ($\theta^a \prec \theta^b$) if and only if $\mathcal{L}_i(\theta^a) \leq \mathcal{L}_i(\theta^b), \forall i \in \{1,...,m\}$ and $\mathcal{L}_j(\theta^a) < \mathcal{L}_j(\theta^b), \exists j \in \{1,...,m\}$.

\textbf{Pareto optimality.} $\theta^{\ast}$ is a Pareto optimal point and $\mathcal{L}(\theta^{\ast})$ is a Pareto optimal objective vector if it does not exist $\hat \theta \in  \Omega$ such that $\hat \theta \prec \theta^{\ast}$.
The set of all Pareto optimal points is called the Pareto set. The image of the Pareto set in the loss space is called the Pareto front.

In this paper, we focus on finding a set of well-representative Pareto solutions that can approximate the Pareto front. This idea and the comparison results of our proposed method with two others are presented in Fig.~\ref{PF_toy_example}.

\subsection{Linear Scalarization}

Linear scalarization is the most commonly-used approach for solving multi-task learning problems. This approach uses a linear weighted sum method to combine the losses of all tasks into a single surrogate loss:	
\begin{eqnarray}\label{eq:sum}
\mathrm{min}_{\theta} \ \mathcal{L}(\theta) = \sum_{i=1}^{m}\vw_i \mathcal{L}_i(\theta),
\end{eqnarray}
where $\vw_i$ is the weight for the $i$-th task. This approach is simple and straightforward, but it has some drawbacks from both multi-task learning and multi-objective optimization perspectives.

In a typical multi-task learning application, the weights $\vw_i$ are needed to be assigned manually before optimization, and the overall performance is highly dependent on the assigned weights. Choosing a proper weight vector could be very difficult even for an experienced MTL practitioner who has expertise on the given problem.

Solving a set of linear scalarization problems with different weight assignments is also not a good idea for multi-objective optimization. As pointed out in~\cite[Chapter~4.7]{boyd2004convex}, this method can only provide solutions on the convex part of the Pareto front. The linear scalarization method with different weight assignments is unable to handle a concave Pareto front as shown in Fig.~\ref{PF_toy_example}.

\subsection{Gradient-based method for multi-objective optimization}

Many gradient-based methods have been proposed for solving multi-objective optimization problems~\cite{desideri2012mutiple,fliege2016method}. Fliege and Svaiter~\cite{fliege2000steepest} have proposed a simple gradient-based method, which is a generalization of a single objective steepest descent algorithm. The update rule of the algorithm is $\theta_{t + 1} = \theta_t + \eta d_t$ where $\eta$ is the step size and the search direction $d_t$ is obtained as follows:
\begin{eqnarray}
    \label{submop1}
         (d_t,\alpha_t) = \text{arg} \min_{d\in R^n,\alpha \in R} \alpha + \frac{1}{2} \norm{d}^2, s.t.~~~ \nabla \mathcal{L}_i(\theta_t)^Td \leq \alpha,  i = 1,...,m.
\end{eqnarray}
The solutions of the above problem will satisfy:

\textbf{Lemma 1~\cite{fliege2000steepest}:} Let $(d^k,\alpha^k)$ be the solution of problem~(\ref{submop1}).
\begin{enumerate}
  \item If $\theta_t$ is Pareto critical, then $d_t = 0 \in \mathbb{R}^n$ and $\alpha_t = 0$.
  \item If $\theta_t$ is not Pareto critical, then
    \begin{eqnarray}\label{eq:bound}
    \begin{aligned}
    & \alpha_t \leq -(1/2) \norm{d_t}^2 < 0, \\
    & \nabla \mathcal{L}_i(\theta_t)^Td_t \leq \alpha_t, i = 1,...,m,
    \end{aligned}
    \end{eqnarray}
\end{enumerate}
where $\theta$ is called Pareto critical if no other solution in its neighborhood can have better values in all objective functions. In other words, if $d_t = \boldsymbol{0}$, no direction can improve the performance for all tasks at the same time. If we want to improve the performance for a specific task, another task's performance will be deteriorated (e.g., $\exists i, \mathcal{L}_i(\theta_t)^Td_t > 0$). Therefore, the current solution is a Pareto critical point. When $d_t \neq \boldsymbol{0}$, we have $\nabla \mathcal{L}_i(\theta_t)^Td_t < 0, i = 1,...,m$, which means $d_t$ is a valid descent direction for all tasks. The current solution should be updated along the obtained direction $\theta_{t + 1} = \theta_t + \eta d_t$.

Recently, Sener and Koltun~\cite{sener2018multi} used the multiple gradient descent algorithm (MGDA)~\cite{desideri2012mutiple} for solving MTL problems and achieve promising results. However, this method does not have a systemic way to incorporate different trade-off preference information. As shown in Fig.~\ref{PF_toy_example}, running the algorithm multiple times can only generate some solutions in the middle of the Pareto front on the synthetic example. In this paper, we generalize this method and propose a novel Pareto MTL algorithm to find a set of well-distributed Pareto solutions with different trade-offs among all tasks.

\section{Pareto Multi-Task Learning}

\subsection{MTL Decomposition}

\begin{wrapfigure}{R}{0.45\linewidth}
	\begin{minipage}{\linewidth}
    \begin{figure}[H]
    	\centering
    	\includegraphics[width=1 \linewidth]{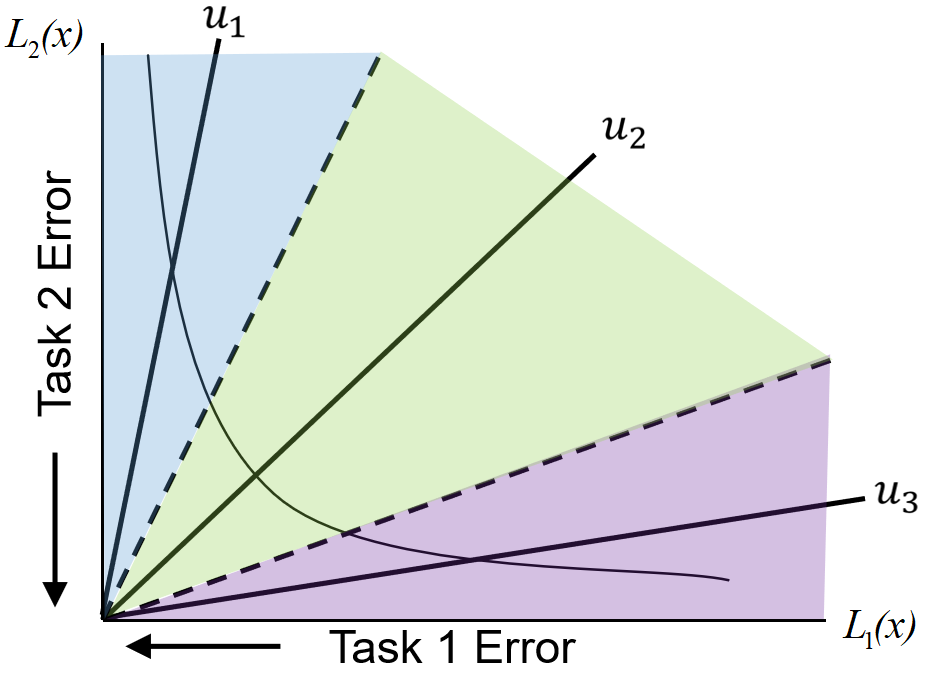}
    	\caption{\textbf{Pareto MTL} decomposes a given MTL problem into several subproblems with a set of preference vectors. Each MTL subproblem aims at finding one Pareto solution in its restricted preference region.}
        \label{MOPM2M}
    \end{figure}
	\end{minipage}
\end{wrapfigure}

We propose the Pareto Multi-Task Learning (Pareto MTL) algorithm in this section. The main idea of Pareto MTL is to decompose a MTL problem into several constrained multi-objective subproblems with different trade-off preferences among the tasks in the original MTL. By solving these subproblems in parallel, a MTL practitioner can obtain a set of well-representative solutions with different trade-offs.

Decomposition-based multi-objective evolutionary algorithm~\cite{zhang2007moea,trivedi2016survey}, which decomposes a multi-objective optimization problem (MOP) into several subproblems and solves them simultaneously, is one of the most popular gradient-free multi-objective optimization methods. Our proposed Pareto MTL algorithm generalizes the decomposition idea for solving large-scale and gradient-based MTL.

We adopt the idea from~\cite{liu2014decomposition} and decompose the MTL into $K$ subproblems with a set of well-distributed unit preference vectors $\{\vu_1,\vu_2,...,\vu_K\}$ in $R^m_{+}$. Suppose all objectives in the MOP are non-negative, the multi-objective subproblem corresponding to the preference vector $\vu_k$ is:

\begin{eqnarray}
    \label{submop}
       \min_{\theta}  \mathcal{L}(\theta) = (\mathcal{L}_1(\theta),\mathcal{L}_2(\theta),\cdots,
\mathcal{L}_m(\theta))^{\mathrm{T}},   s.t.~~~ \mathcal{L}(\theta) \in \Omega_k,
\end{eqnarray}
where $\Omega_k (k = 1,...,K)$ is a subregion in the objective space:
\begin{eqnarray}
    \label{subregion}
	\Omega_k = \{\vv \in R^m_{+}| \vu_j^T\vv \leq \vu_k^T\vv, \forall j = 1,...,K  \}
\end{eqnarray}
and $\vu_j^T\vv$ is the inner product between the preference vector $\vu_j$ and a given vector $\vv$. That is to say, $\vv \in \Omega_k$  if and only if $\vv$ has the smallest acute angle to $\vu_k$ and hence the largest inner product $\vu_k^T\vv$ among all $K$ preference vectors.

The subproblem (\ref{submop}) can be further reformulated as:
\begin{eqnarray}
    \label{submop_v1}
    \begin{aligned}
        & \min_{\theta} \mathcal{L}(\theta) = (\mathcal{L}_1(\theta),\mathcal{L}_2(\theta),\cdots,
\mathcal{L}_m(\theta))^{\mathrm{T}}\\
        &s.t.~~~ \mathcal{G}_j(\theta_t) = (  \vu_j -\vu_k)^T \mathcal{L}(\theta_t) \leq 0, \forall j = 1,...,K, \\
    \end{aligned}
\end{eqnarray}

As shown in Fig.~\ref{MOPM2M}, the preference vectors divide the objective space into different sub-regions. The solution for each subproblem would be attracted by the corresponding preference vector and hence be guided to its representative sub-region. The set of solutions for all subproblems would be in different sub-regions and represent different trade-offs among the tasks.

\subsection{Gradient-based Method for Solving Subproblems}

\subsubsection{Finding the Initial Solution}
To solve the constrained multi-objective subproblem (\ref{submop}) with a gradient-based method, we need to find an initial solution which is feasible or at least satisfies most constraints. For a randomly generated solution $\theta_r$, one straightforward method is to find a feasible initial solution $\theta_0$ which satisfies:
\begin{eqnarray}
    \label{init_sol}
    \begin{aligned}
        & \min_{\theta_0} \norm{\theta_0 - \theta_r}^2
        &s.t.~~~ \mathcal{L}(\theta_0) \in \Omega_k.
    \end{aligned}
\end{eqnarray}
However, this projection approach is an $n$ dimensional constrained optimization problem~\cite{gebken2017descent}. It is inefficient to solve this problem directly, especially for a deep neural network with millions of parameters. In the proposed Pareto MTL algorithm, we reformulate this problem as unconstrained optimization, and use a sequential gradient-based method to find the initial solution $\theta_0$.

For a solution $\theta_{r}$, we define the index set of all activated constraints as $I(\theta_{r}) = \{j |\mathcal{G}_j(\theta_{r}) \geq 0, j = 1,...,K \}$. We can find a valid descent direction $d_{r}$ to reduce the value of all activated constraints $\{\mathcal{G}_j(\theta_{r})|j \in I(\theta_{r})\}$ by solving:
\begin{eqnarray}
    \label{submop1_feasible}
         (d_{r},\alpha_{r}) = \text{arg} \min_{d\in R^n,\alpha \in R} \alpha + \frac{1}{2} \norm{d}^2, s.t.\nabla \mathcal{G}_j(\theta_{r})^Td \leq \alpha, j \in I(\theta_{r}).
\end{eqnarray}
This approach is similar to the unconstrained gradient-based method (\ref{submop1}), but it reduces the value of all activated constraints. The gradient-based update rule is $\theta_{r_{t+1}} = \theta_{r_t} + \eta_r d_{r_t}$ and will be stopped once a feasible solution is found  or a predefined number of iterations is met.

\subsubsection{Solving the Subproblem}
Once we have an initial solution, we can use a gradient-based method to solve the constrained subproblem. For a constrained multiobjective optimization problem, the Pareto optimality restricted on the feasible region $\Omega_k$ can be defined as~\cite{fliege2000steepest}:

\textbf{Restricted Pareto Optimality.} $\theta^{\ast}$ is a Pareto optimal point for $\mathcal{L}(\theta)$ restricted on $\Omega_k$ if $\theta^{\ast} \in \Omega_k$ and  it does not exist $\hat \theta \in  \Omega_k$ such that $\hat \theta \prec \theta^{\ast}$.

According to~\cite{fliege2000steepest,gebken2017descent}, we can find a descent direction for this constrained MOP by solving a subproblem similar to the subproblem ({\ref{submop1}}) for the unconstrained case:
\begin{eqnarray}
    \label{submop2}
    \begin{aligned}
         (d_t,\alpha_t) = &\text{arg} \min_{d\in R^n,\alpha \in R} \alpha + \frac{1}{2} \norm{d}^2 \\
        & s.t.~~~ \nabla \mathcal{L}_i(\theta_t)^Td \leq \alpha, i = 1,...,m. \\
         & ~~~~~~~~ \nabla \mathcal{G}_j(\theta_t)^Td \leq \alpha, j \in I_{\epsilon}(\theta_t),
    \end{aligned}
\end{eqnarray}
where $I_{\epsilon}(\theta)$ is the index set of all activated constraints:
\begin{eqnarray}
    \label{act_set}
    I_{\epsilon}(\theta) = \{j \in I|\mathcal{G}_j(\theta) \geq -\epsilon \}.
\end{eqnarray}

We add a small threshold $\epsilon$ to deal with the solutions near the constraint boundary. Similar to the unconstrained case, for a feasible solution $\theta_t$, by solving problem~(\ref{submop2}), we either obtain $d_t = \boldsymbol{0}$ and confirm that $\theta_t$ is a Pareto critical point restricted on $\Omega_k$, or obtain $d_t \neq \boldsymbol{0}$ as a descent direction for the constrained multi-objective problem~(\ref{submop_v1}). In the latter case, if all constraints are inactivated (e.g., $I_{\epsilon}(\theta) = \emptyset$), $d_t$ is a valid descent direction for all tasks. Otherwise, $d_t$ is a valid direction to reduce the values for all tasks and all activated constraints.

\textbf{Lemma 2~\cite{gebken2017descent}:} Let $(d^k,\alpha^k)$ be the solution of problem~(\ref{submop2}).
\begin{enumerate}
  \item If $\theta_t$ is Pareto critical restricted on $\Omega_k$, then $d_t = \boldsymbol{0} \in \mathbb{R}^n$ and $\alpha_t = 0$.
  \item If $\theta_t$ is not Pareto critical restricted on $\Omega_k$, then
    \begin{eqnarray}\label{eq:bound_v2}
    \begin{aligned}
    & \alpha_t \leq -(1/2) \norm{d_t}^2 < 0, \\
    & \nabla \mathcal{L}_i(\theta_t)^Td_t \leq \alpha_t, i = 1,...,m \\
    & \nabla \mathcal{G}_j(\theta_t)^Td_t \leq \alpha_t, j \in I_{\epsilon}(\theta_t).
    \end{aligned}
    \end{eqnarray}
\end{enumerate}

Therefore, we can obtain a restricted Pareto critical solution for each subproblem with simple iterative gradient-based update rule $\theta_{t+1} = \theta_{t} + \eta_r d_t$. By solving all subproblems, we can obtain a set of diverse Pareto critical solutions restricted on different sub-regions, which can represent different trade-offs among all tasks for the original MTL problem.

\subsubsection{Scalable Optimization Method}
By solving the constrained optimization problem~(\ref{submop2}), we can obtain a valid descent direction for each multi-objective constrained subproblem. However, the optimization problem itself is not scalable well for high dimensional decision space. For example, when training a deep neural network, we often have more than millions of parameters to be optimized, and solving the constrained optimization problem~(\ref{submop2}) in this scale would be extremely slow. In this subsection, we propose a scalable optimization method to solve the constrained optimization problem.

Inspired by~\cite{fliege2000steepest}, we first rewrite the optimization problem~(\ref{submop2}) in its dual form. Based on the KKT conditions, we have
\begin{eqnarray}
    \label{submop2_KKT}
         d_t = - (\sum_{i=1}^{m} \lambda_i \nabla \mathcal{L}_i(\theta_t) +  \sum_{j \in I_{\epsilon(\vx)}} \beta_i \nabla \mathcal{G}_j(\theta_t)), ~~\sum_{i=1}^{m} \lambda_i + \sum_{j \in I_{\epsilon(\theta)}} \beta_j = 1,
\end{eqnarray}
where $\lambda_i \geq 0$ and $\beta_i \geq 0$ are the Lagrange multipliers for the linear inequality constraints. Therefore, the dual problem is:
\begin{eqnarray}
    \label{submop2_dual}
    \begin{aligned}
        &\max_{\lambda_i,\beta_j}~ -\frac{1}{2}\norm{\sum_{i=1}^{m} \lambda_i \nabla \mathcal{L}_i(\theta_t) +  \sum_{j \in I_{\epsilon(\vx)}} \beta_i \nabla \mathcal{G}_j(\theta_t)}^2 \\
        & s.t.~~~ \sum_{i=1}^{m} \lambda_i + \sum_{j \in I_{\epsilon(\theta)}} \beta_j = 1,  \lambda_i \geq 0, \beta_j \geq 0, \forall i = 1,...,m, \forall j \in I_{\epsilon}(\theta).
    \end{aligned}
\end{eqnarray}
For the above problem, the decision space is no longer the parameter space, and it becomes the objective and constraint space. For a multiobjective optimization problem with $2$ objective function and $5$ activated constraints, the dimension of problem~(\ref{submop2_dual}) is $7$, which is significantly smaller than the dimension of problem~(\ref{submop2}) which could be more than a million.

The algorithm framework of Pareto MTL is shown in \textbf{Algorithm}~\ref{alg:ParetoMTL}. All subproblems can be solved in parallel since there is no communication between them during the optimization process. The only preference information for each subproblem is the set of preference vectors. Without any prior knowledge for the MTL problem, a set of evenly distributed unit preference vectors would be a reasonable default choice, such as $K+1$ preference vectors $\{(cos(\frac{k\pi}{2K}),sin(\frac{k\pi}{2K}))| k = 0,1,...,K\}$ for 2 tasks. We provide more discussion on preference vector setting and sensitivity analysis of the preference vectors in the supplementary material.

\clearpage

\begin{algorithm}[H]
	\caption{Pareto MTL Algorithm}
	\label{alg:ParetoMTL}
	\begin{algorithmic}[1]
		
		\STATE \textbf{Input:} A set of evenly distributed vectors $\{\vu_1,\vu_2,...,\vu_K\}$
		\STATE \textbf{Update Rule:}
		\STATE (can be solved in parallel)
		\FOR{$k = 1$ to $K$}
		    \STATE randomly generate parameters $\theta_r^{(k)}$
		    \STATE find the initial parameters $\theta_0^{(k)}$ from $\theta_r^{(k)}$ using gradient-based method
    		\FOR{$t = 1$ to $T$}

        		\STATE obtain $\lambda_{ti}^{(k)} \geq 0, \beta_{ti}^{(k)} \geq 0, \forall i = 1,...,m, \forall j \in I_{\epsilon}(\theta)$ by solving subproblem~(\ref{submop2_dual})
        		\STATE calculate the direction $d_t^{(k)} = - (\sum_{i=1}^{m} \lambda_{ti}^{(k)} \nabla \mathcal{L}_i(\theta^{(k)}_{t}) +  \sum_{j \in I_{\epsilon(\vx)}} \beta_{ti}^{(k)} \nabla \mathcal{G}_j(\theta^{(k)}_{t})$
        		\STATE update the parameters $\theta^{(k)}_{t + 1} = \theta^{(k)}_{t} + \eta d_t^{(k)}$
    		\ENDFOR
		\ENDFOR	
		\STATE \textbf{Output:} The set of solutions for all subproblems with different trade-offs  $\{\theta^{(k)}_{T}| k = 1,\cdot,K\} $
		
	\end{algorithmic}
\end{algorithm}

\subsection{Pareto MTL as an Adaptive Linear Scalarization Approach}

We have proposed the Pareto MTL algorithm from the multi-objective optimization perspective. In this subsection, we show that the Pareto MTL algorithm can be reformulated as a linear scalarization of tasks with adaptive weight assignment. In this way, we can have a deeper understanding of the differences between Pareto MTL and other MTL algorithms.

We first tackle the unconstrained case. Suppose we do not decompose the multi-objective problem and hence remove all constraints from the problem (\ref{submop2_dual}), it will immediately reduce to the update rule proposed by MGDA~\cite{desideri2012mutiple} which is used in~\cite{sener2018multi}. It is straightforward to rewrite the corresponding MTL into a linear scalarization form:
\begin{eqnarray}
    \label{update_mtl_nc}
    \begin{aligned}
        & \mathcal{L}(\theta_t) = \sum_{i=1}^{m} \lambda_i  \mathcal{L}_i(\theta_t),
    \end{aligned}
\end{eqnarray}
where we adaptively assign the weights $\lambda_i$ by solving the following problem in each iteration:
\begin{eqnarray}
    \label{submop2_dual_nc}
        \max_{\lambda_i} -\frac{1}{2}\norm{\sum_{i=1}^{m} \lambda_i \nabla \mathcal{L}_i(\theta_t)}^2, ~~~
         s.t.~ \sum_{i=1}^{m} \lambda_i = 1, ~~~ \lambda_i \geq 0, \forall i = 1,...,m.
\end{eqnarray}

In the constrained case, we have extra constraint terms $ \mathcal{G}_j(\theta_t)$.  If $\mathcal{G}_j(\theta_t)$ is inactivated, we can ignore it. For an activated $\mathcal{G}_j(\theta_t)$, assuming the corresponding reference vector is $\vu_k$, we have:
\begin{eqnarray}
    \label{g_act}
    \nabla \mathcal{G}_j(\theta_t)  = (  \vu_j -\vu_k)^T \nabla \mathcal{L}(\theta_t) = \sum_{i=1}^{m} (\vu_{ji} - \vu_{ki}) \nabla \mathcal{L}_i(\theta_t).
\end{eqnarray}
Since the gradient direction $d_t$ can be written as a linear combination of all $\nabla \mathcal{L}_i(\theta_t)$ and $\nabla \mathcal{G}_j(\theta_t)$ as in~(\ref{submop2_KKT}), the general Pareto MTL algorithm can be rewritten as:
\begin{eqnarray}
    \label{update_mtl_c}
    \mathcal{L}(\theta_t) = \sum_{i=1}^{m} \alpha_i  \mathcal{L}_i(\theta_t), ~\text{where }  \alpha_i = \lambda_i + \sum_{j \in I_{\epsilon(\theta)}} \beta_j (\vu_{ji} - \vu_{ki}),
\end{eqnarray}
where $\lambda_i$ and $\beta_j$ are obtained by solving problem (\ref{submop2_dual}) with assigned reference vector $\vu_{k}$.

Therefore, although MOO-MTL~\cite{sener2018multi} and Pareto MTL are both derived from multi-objective optimization, they can also be treated as linear MTL scalarization with adaptive weight assignments. Both methods are orthogonal to many existing MTL approaches. We provide further discussion on the adaptive weight vectors in the supplementary material.

\section{A Synthetic Example}

To better analyze the convergence behavior of the proposed Pareto MTL, we first compare it with two commonly used methods, namely the linear scalarization method and the multiple gradient descent algorithm used in MOO-MTL~\cite{sener2018multi}, on a simple synthetic multi-objective optimization problem:
\begin{eqnarray}
    \label{toy_example}
    \begin{aligned}
        &\min_{\vx} f_1(\vx) = 1 - \exp{(- \sum_{i=1}^{d} (\vx_d - \frac{1}{\sqrt{d}})^2)} \\
        &\min_{\vx} f_2(\vx) = 1 - \exp{(- \sum_{i=1}^{d} (\vx_d + \frac{1}{\sqrt{d}})^2)}
    \end{aligned}
\end{eqnarray}
where $f_1(\vx)$ and $f_2(\vx)$ are two objective functions to be minimized at the same time and $\vx = (\vx_1,\vx_2,...,\vx_d)$ is the $d$ dimensional decision variable. This problem has a concave Pareto front on the objective space.

The results obtained by different algorithms are shown in Fig.~\ref{PF_toy_example}. In this case, the proposed Pareto MTL can successfully find a set of well-distributed Pareto solutions with different trade-offs. Since MOO-MTL tries to balance different tasks during the optimization process, it gets a set of solutions with similar trade-offs in the middle of the Pareto front in multiple runs. It is also interesting to observe that the linear scalarization method can only generate extreme solutions for the concave Pareto front evenly with 100 runs. This observation is consistent with the theoretical analysis in~\cite{boyd2004convex} that the linear scalarization method will miss all concave parts of a Pareto front. It is evident that fixed linear scalarization is not always a good idea for solving the MTL problem from the multi-objective optimization perspective.

\section{Experiments}

In this section, we compare our proposed Pareto MTL algorithm on different MTL problems with the following algorithms: 1) \textbf{Single Task}: the single task baseline; 2) \textbf{Grid Search}: linear scalarization with fixed weights; 3) \textbf{GradNorm}~\cite{chen2018grad}:  gradient normalization; 4) \textbf{Uncertainty}~\cite{kendall2017multi}: adaptive weight assignments with uncertainty balance; 5) \textbf{MOO-MTL}~\cite{sener2018multi}: finding one Pareto optimal solution for multi-objective optimization problem. More experimental results and discussion can be found in the supplementary material.

\subsection{Multi-Fashion-MNIST}

\begin{figure}[H]
\centering
\subfloat[MultiMNIST]{\includegraphics[width = 0.33\textwidth]{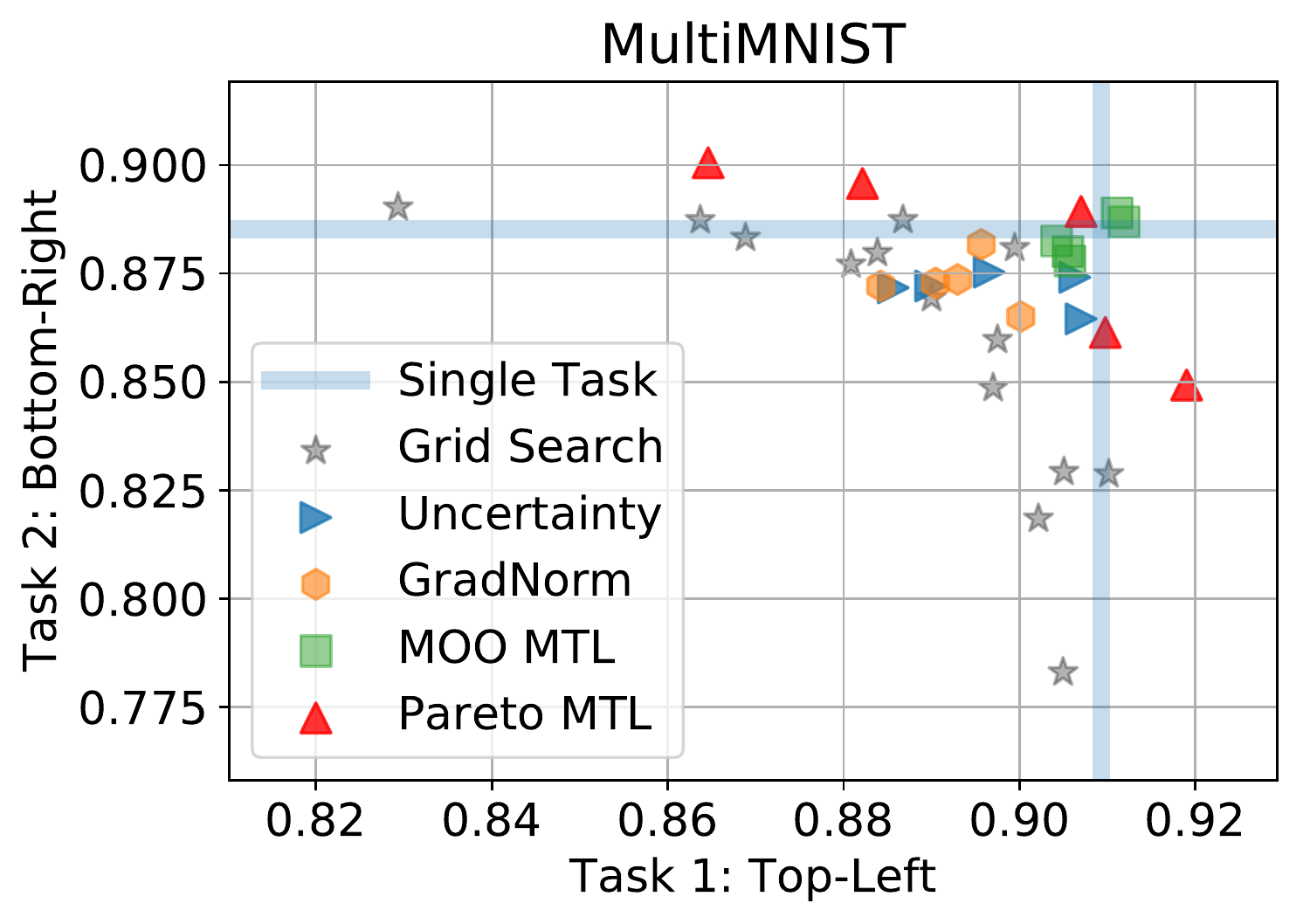}}
\subfloat[MultiFashionMNIST]{\includegraphics[width = 0.33\textwidth]{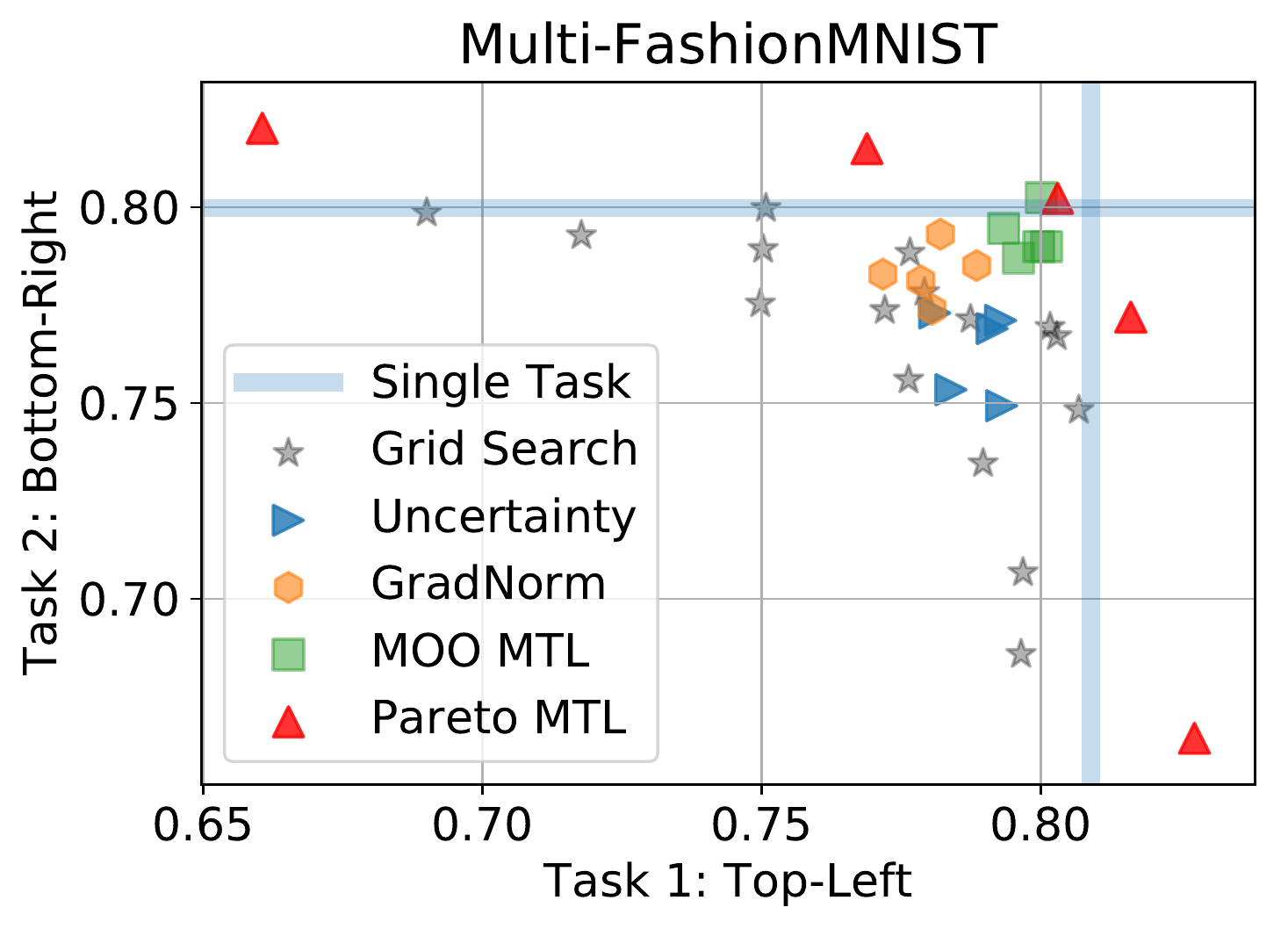}}
\subfloat[Multi-(Fashion+MNIST)]{\includegraphics[width = 0.33\textwidth]{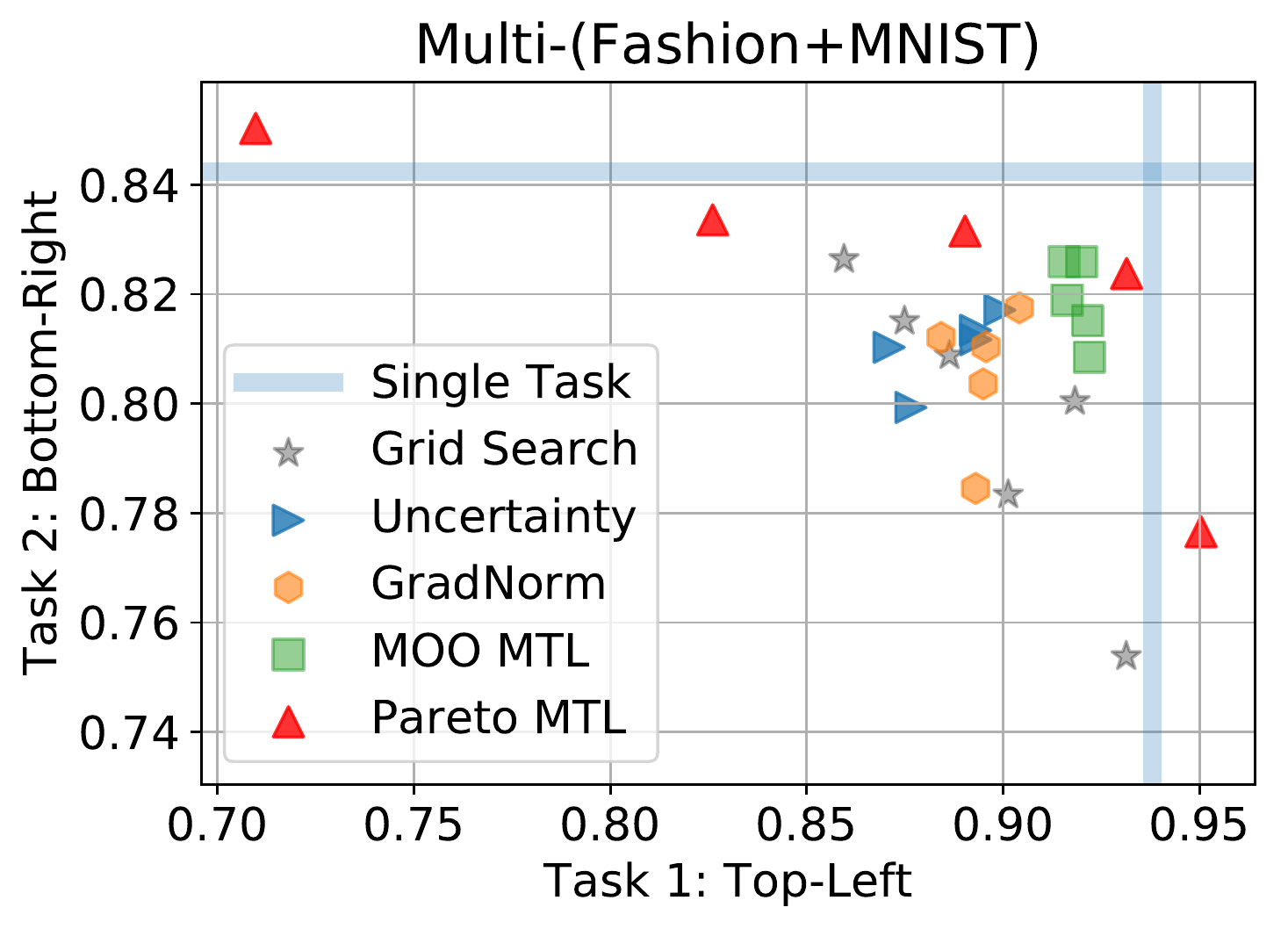}} \\
\caption{\textbf{The results for three experiments with Task1\&2 accuracy:} our proposed Pareto MTL can successfully find a set of well-distributed solutions with different trade-offs for all experiments, and it significantly outperforms Grid Search, Uncertainty and GradNorm. MOO-MTL algorithm can also find promising solutions, but their diversity is worse than the solutions generated by Pareto MTL. }

\label{mnist_results}
\end{figure}

In order to evaluate the performance of Pareto MTL on multi-task learning problems with different tasks relations, we first conduct experiments on MultiMNIST~\cite{sabour2017dynamic} and two MultiMNIST-like datasets. To construct the MultiMNIST dataset, we randomly pick two images with different digits from the original MNIST dataset~\cite{lecun1998gradient}, and then combine these two images into a new one by putting one digit on the top-left corner and the other one on the bottom-right corner. Each digit can be moved up to $4$ pixels on each direction. With the same approach, we can construct a MultiFashionMINST dataset with overlap FashionMNIST items~\cite{xiao2017fashion}, and a Multi-(Fashion + MNIST) with overlap MNIST and FashionMNIST items.  For each dataset, we have a two objective MTL problem to classify the item on the top-left (task 1) and to classify the item on the bottom-right (task 2). We build a LeNet~\cite{lecun1998gradient} based MTL neural network similar to the one used in~\cite{sener2018multi}. The obtained results are shown in Fig.~\ref{mnist_results}.

In all experiments, since the tasks conflict with each other, solving each task separately results in a hard-to-beat single-task baseline. Our proposed Pareto MTL algorithm can generate multiple well-distributed Pareto solutions for all experiments, which are compatible with the strong single-task baseline but with different trade-offs among the tasks. Pareto MTL algorithm achieves the overall best performance among all MTL algorithms. These results confirm that our proposed Pareto MTL can successfully provide a set of well-representative Pareto solutions for a MTL problem.

It is not surprising to observe that the Pareto MTL's solution for subproblems with extreme preference vectors (e.g., $(0,1)$ and $(1,0)$) always have the best performance in the corresponding task. Especially in the Multi-(Fashion-MNIST) experiment, where the two tasks are less correlated with each other. In this problem, almost all MTL's solutions are dominated by the strong single task's baseline. However, Pareto MTL can still generate solutions with the best performance for each task separately. It behaves as auxiliary learning, where the task with the assigned preference vector is the main task, and the others are auxiliary tasks.

Pareto MTL uses neural networks with simple hard parameter sharing architectures as the base model for MTL problems. It will be very interesting to generalize Pareto MTL to other advanced soft parameter sharing architectures~\cite{ruder2017overview}. Some recently proposed works on task relation learning~\cite{zamir2018taskonomy,Ma2018Modeling,zhang2018learning} could also be useful for Pareto MTL to make better trade-offs for less relevant tasks.

\subsection{Self-Driving Car: Localization}

\begin{figure}[H]
    \centering
    \subfloat{
        \adjustbox{valign=b}{

        \begin{tabular}{cccc}
            \hline
            Method      & Reference             & Translation    & Rotation                                 \\
                        & Vector                & (m)            & ($^\circ$) \\ \hline
            Single Task & \textbf{-}            & 8.392          & 2.461                                    \\ \hline
                        & (0.25,0.75)           & 9.927          & 2.177                                    \\
            Grid Search & (0.5,0.5)             & 7.840          & 2.306                                    \\
                        & (0.75,0.25)           & 7.585          & 2.621                                    \\ \hline
            GradNorm    & -                     & 7.751          & 2.287                                    \\
            Uncertainty & -                     & 7.624          & 2.263                                    \\
            MOO-MTL     & -                     & 7.909          & 2.090                                    \\ \hline
                        & (0,1)                 & \textbf{7.285} & 2.335                                    \\
            Pareto MTL  & ($\frac{\sqrt{2}}{2},\frac{\sqrt{2}}{2}$) & 7.724          & 2.156                                    \\
                        & (1,0)                 & 8.411          & \textbf{1.771}                           \\ \hline
        \end{tabular}
       }
    }
    \subfloat{

        \includegraphics[width=0.4\linewidth,valign=b]{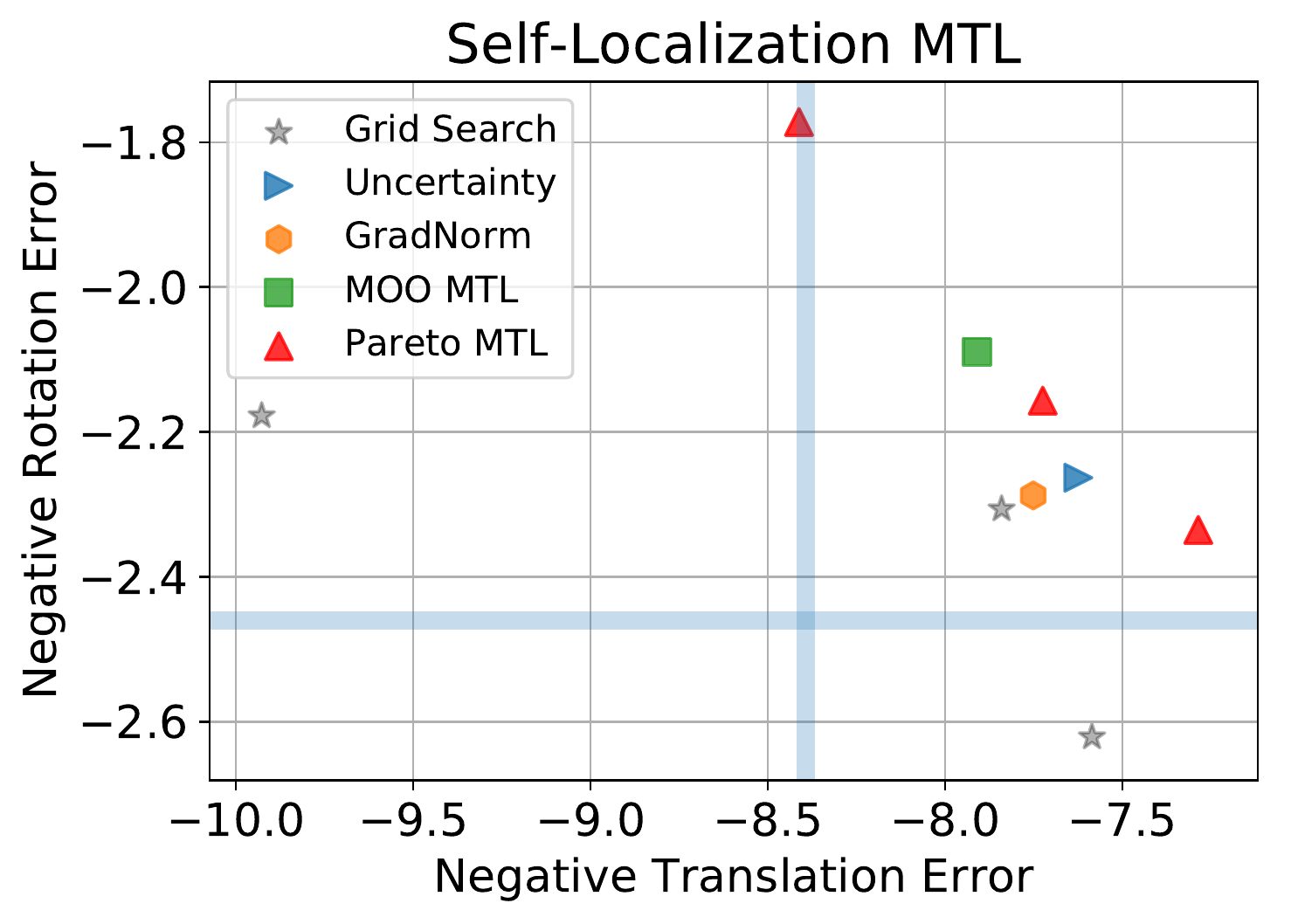}

        }
    \caption{\textbf{The results of self-location MTL experiment:}  Our proposed Pareto MTL outperforms other algorithms and provides solutions with different trade-offs.}
\label{PF_self_localization}
\end{figure}

We further evaluate Pareto MTL on an autonomous driving self-localization problem~\cite{wang2018dels}. In this experiment, we simultaneously estimate the location and orientation of a camera put on a driving car based on the images it takes. We use data from the apolloscape autonomous driving dataset~\cite{apolloscape_arXiv_2018,wang2019the}, and focus on the Zpark sample subset. We build a PoseNet with a ResNet18~\cite{kendall2015posenet} encoder as the MTL model. The experiment results are shown in Fig.~\ref{PF_self_localization}. It is obvious that our proposed Pareto MTL can generate solutions with different trade-offs and outperform other MTL approaches.

We provide more experiment results and analysis on finding the initial solution, Pareto MTL with many tasks, and other relative discussions in the supplementary material.

\section{Conclusion}
In this paper, we proposed a novel Pareto Multi-Task Learning (Pareto MTL) algorithm to generate a set of well-distributed Pareto solutions with different trade-offs among tasks for a given multi-task learning (MTL) problem. MTL practitioners can then easily select their preferred solutions among these Pareto solutions. Experimental results confirm that our proposed algorithm outperforms some state-of-the-art MTL algorithms and can successfully find a set of well-representative solutions for different MTL applications.

\clearpage

\subsubsection*{Acknowledgments}
This work was supported by the Natural Science Foundation of China under Grant 61876163 and Grant 61672443, ANR/RGC Joint Research Scheme sponsored by the Research Grants Council of the Hong Kong Special Administrative Region, China and France National Research Agency (Project No. A-CityU101/16), and Hong Kong RGC General Research Funds under Grant 9042489 (CityU 11206317) and Grant 9042322 (CityU 11200116).



\medskip

\bibliographystyle{unsrt}
\bibliography{neurips_2019}

\clearpage

\appendix
\section*{Supplementary Material: Pareto MTL}

\addcontentsline{toc}{section}{Supplementary Material: Pareto MTL}
\renewcommand{\thesubsection}{\Alph{subsection}}

In this supplementary material, we provide more detailed discussions and experimental results on Pareto MTL. We also point out some limitations for the current Pareto MTL and propose some potential research directions.

\section{The Importance of Finding the Initial Solution}

\begin{figure*}[h]
\centering
\subfloat[MOO MTL]{\includegraphics[width = 0.33\textwidth]{Figures/toy_moo_mtl.eps}}
\subfloat[Pareto MTL w/o Initialization]{\includegraphics[width = 0.33\textwidth]{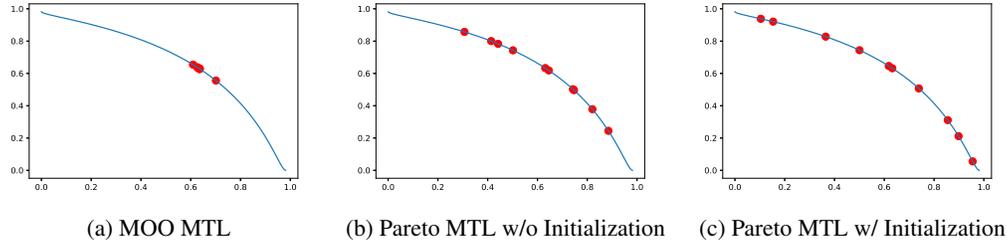}}
\subfloat[Pareto MTL w/ Initialization]{\includegraphics[width = 0.33\textwidth]{Figures/toy_pareto_mtl.eps}}
\caption{\textbf{The convergence behaviours of different algorithms on the synthetic example.} The proposed Pareto MTL algorithm without initialization step can still generate a set of solutions with different trade-offs on the Pareto front. However, it fails to find solutions near the end points.}
\label{PF_noinit}
\end{figure*}

Finding an initial solution is important for solving Pareto MTL. In this paper, we propose a gradient-based method to find an initial solution to each MTL problem. The initial solution should be a feasible solution to the constrained subproblem or at least satisfy most constraints. In this section, we further analyze the importance of finding the initial solution.

We first test the Pareto MTL algorithm without the initialization step on the synthetic example. As shown in Figure.~\ref{PF_noinit}, without the initialization step, Pareto MTL can still generate a set of Pareto solutions with different trade-offs and outperforms the MOO-MTL algorithm. However, the diversity of these solutions is worse than those generated by the Pareto MTL with the initialization step. It is obvious that Pareto MTL without initialization fails to find solutions near the endpoints of the Pareto front. The lack of ability to cross the boundary between different sub-regions would be one reason for inferior performance.

For a subproblem, a randomly generated solution might not be in (or even far away from) its preference sub-region. With the initialization step, the solution can be sequentially updated to get close to the assigned preference sub-region since the constraint values are lowered at each iteration. Many constraints would turn inactivated once the solution crosses its boundary to get close to its assigned preference vector. In contrast, without the initialization step, the solution might stop at some boundary of the preference vectors (with corresponding activated constraints) during the optimization process since now the objective functions are taken into consideration. It is easier to find a descent direction to lower the value of activated constraints than to find a direction to lower both the values of activated constraints and the tasks.

\clearpage

\begin{figure}[H]
	\centering
	\includegraphics[width= 0.50 \linewidth]{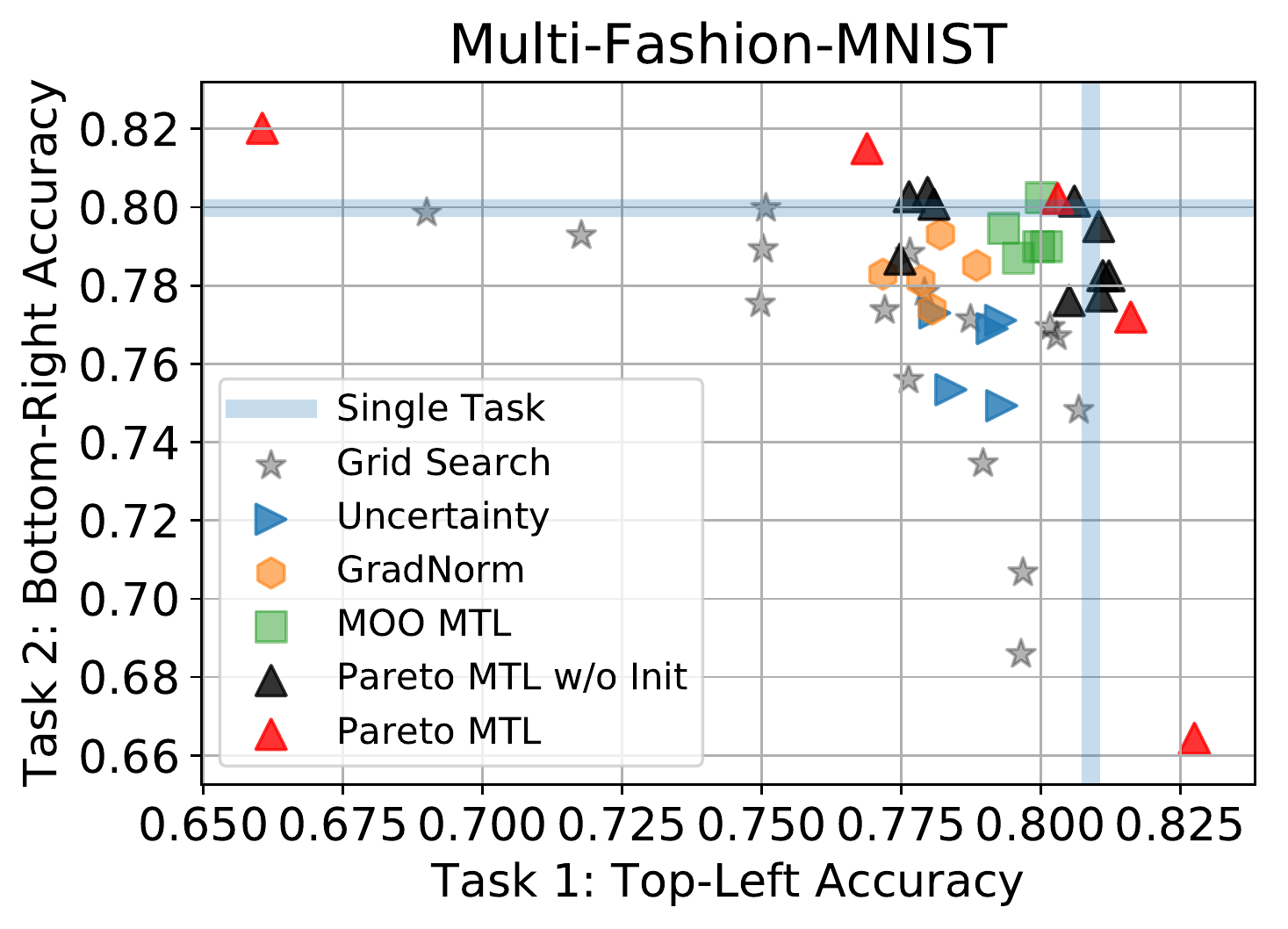}
	\caption{\textbf{Pareto MTL with and without Initialization on Multi-Fashion MNIST problem.}}
    \label{Fashion_two_noinit}
\end{figure}

We also test the Pareto MTL without initialization step on the Multi-FashionMNIST problem. As shown in Fig.~\ref{Fashion_two_noinit}, the Pareto MTL without initialization step can generate solutions with different trade-offs, but the diversity is much worse than Pareto MTL with initialization step.

In the proposed Pareto MTL, we obtain the descent direction for each subproblem by solving:
\begin{eqnarray}
    \label{submop2_app}
    \begin{aligned}
         (d_t,\alpha_t) = &\text{arg} \min_{d\in R^n,\alpha \in R} \alpha + \frac{1}{2} \norm{d}^2 \\
        & s.t.~~~ \nabla \mathcal{L}_i(\theta_t)^Td \leq \alpha, i = 1,...,m. \\
         & ~~~~~~~~ \nabla \mathcal{G}_j(\theta_t)^Td \leq \alpha, j \in I_{\epsilon}(\theta_t).
    \end{aligned}
\end{eqnarray}
Actually, depended on how to obtain the gradient direction, we have three different algorithms:
\begin{enumerate}
  \item Only consider the tasks $\mathcal{L}_i(\theta_t)$: MOO-MTL, for convergence;
  \item Only consider the constraints $\mathcal{G}_j(\theta_t)$: the initialization step in Pareto MTL, for diversity;
  \item Consider both $\mathcal{L}_i(\theta_t)$ and $\mathcal{G}_j(\theta_t)$: the main step of Pareto MTL, tries to find a set of restricted Pareto points on diverse sub-region, somehow balance the convergence and diversity.
\end{enumerate}

How to choose or switch among these different algorithms would be an interesting research topic. The proposed Pareto MTL first runs step 2 and then runs step 3. An immediate extension is to keep and get a snapshot~\cite{huang2017snapshot} of all solutions at the end of step 3, then relax all constraints and run step 1. In this way, we can obtain a set of restricted Pareto critical solutions for each subproblem by Pareto MTL (running step 2 and step 3) with good diversity, plus a set of Pareto critical solutions (not restricted) with potential better convergence by running step 1 at the end.

\section{MTL with Many Tasks}

Pareto MTL uses a set of preference vectors to decompose a given MTL problem into several constrained multi-objective subproblems. By solving all subproblems, Pareto MTL can obtain a set of optimal solutions with different trade-offs among all tasks. However, to fairly cover the whole objective space of all tasks, the number of required preference vectors would increase exponentially when the MTL problem has more tasks.

To be concrete,  Pareto MTL needs to solve a $ (m + K - 1) $-dimensional constrained optimization problem to find the descent direction at each iteration, and solve $K$ subproblems in total, where $m$ is the number of tasks and $K$ is the number of preference vectors. Under mild assumption, the Pareto front would be a $m - 1$ dimensional manifold for a multi-objective optimization problem~\cite{miettinen2012nonlinear}. Suppose we need $p$ (e.g., $5$) widely distributed Pareto solutions to properly represent different optimal trade-offs on one dimension of the Pareto front, the total required solutions could be $p^{m - 1}$ (e.g., $25$ solutions for three tasks and $125$ solutions for four tasks) to cover the whole Pareto front. Since we need to assign one preference vector for each expected solution, the dimension of the constrained optimization problem at each iteration would be $m + p^{m - 1} - 1$, and there are $p^{m - 1}$ MTL subproblems to be solved in total. The current Pareto MTL suffers the curse of dimensionality to cover the whole Pareto front for a MTL with many tasks. In other words, it would be extremely time-consuming for Pareto MTL to provide a set of widely distributed solutions to explore the whole objective space for a MTL problem with many tasks.

To check the required number of solutions, we test Pareto MTL on a multi-task learning problem with three prediction tasks on the UCI census-income dataset~\cite{kohavi1996scaling,Dua2019uci}. This dataset is a subset of the 1994 American Census dataset and contains $299,285$ adults' demographic information records with $40$ different features. Similar to the setting in~\cite{Ma2018Modeling}, we set the income, education level, and marital status as three binary targets to be predicted:

\begin{itemize}
    \item Task 1: whether the person's income exceeds \$50K/year.
    \item Task 2: whether the person's education level is at least college.
    \item Task 3: whether the person is never married.
\end{itemize}

We build a multi-task neural network with hard parameter sharing for three tasks as the prediction model. We first convert all discrete categorical features into one-hot vectors and obtain a $482$ dimensional input feature vector for each record. The model has two hidden fully connected hidden layers with $1024$ and $128$ hidden units, and each task has its own output layer.

\begin{figure}[H]
\centering
\subfloat[Pareto MTL with $25$ and $5$ preference vectors]{\includegraphics[width = 0.50\textwidth]{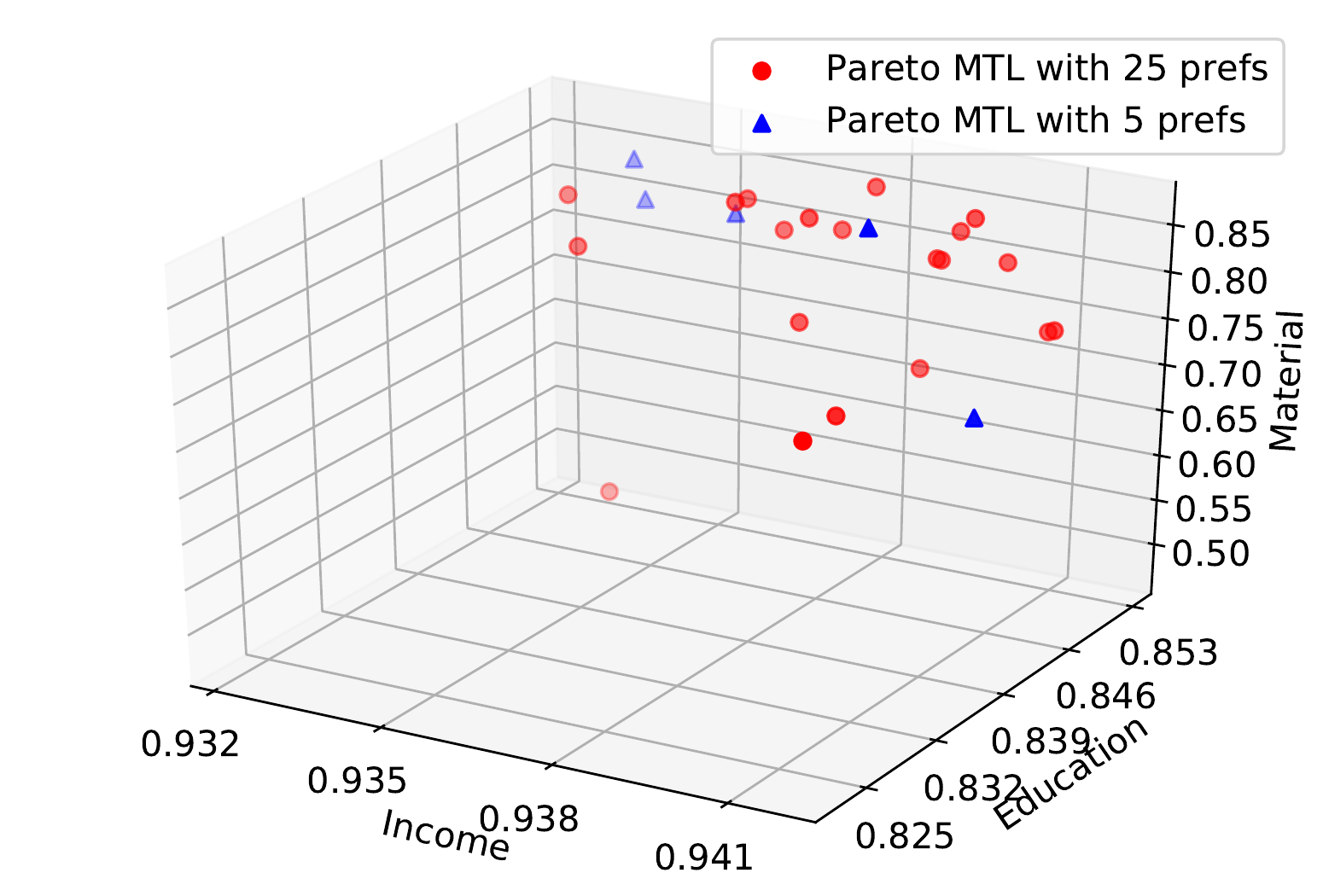}}
\subfloat[Pareto MTL v.s. Random Search]{\includegraphics[width = 0.50\textwidth]{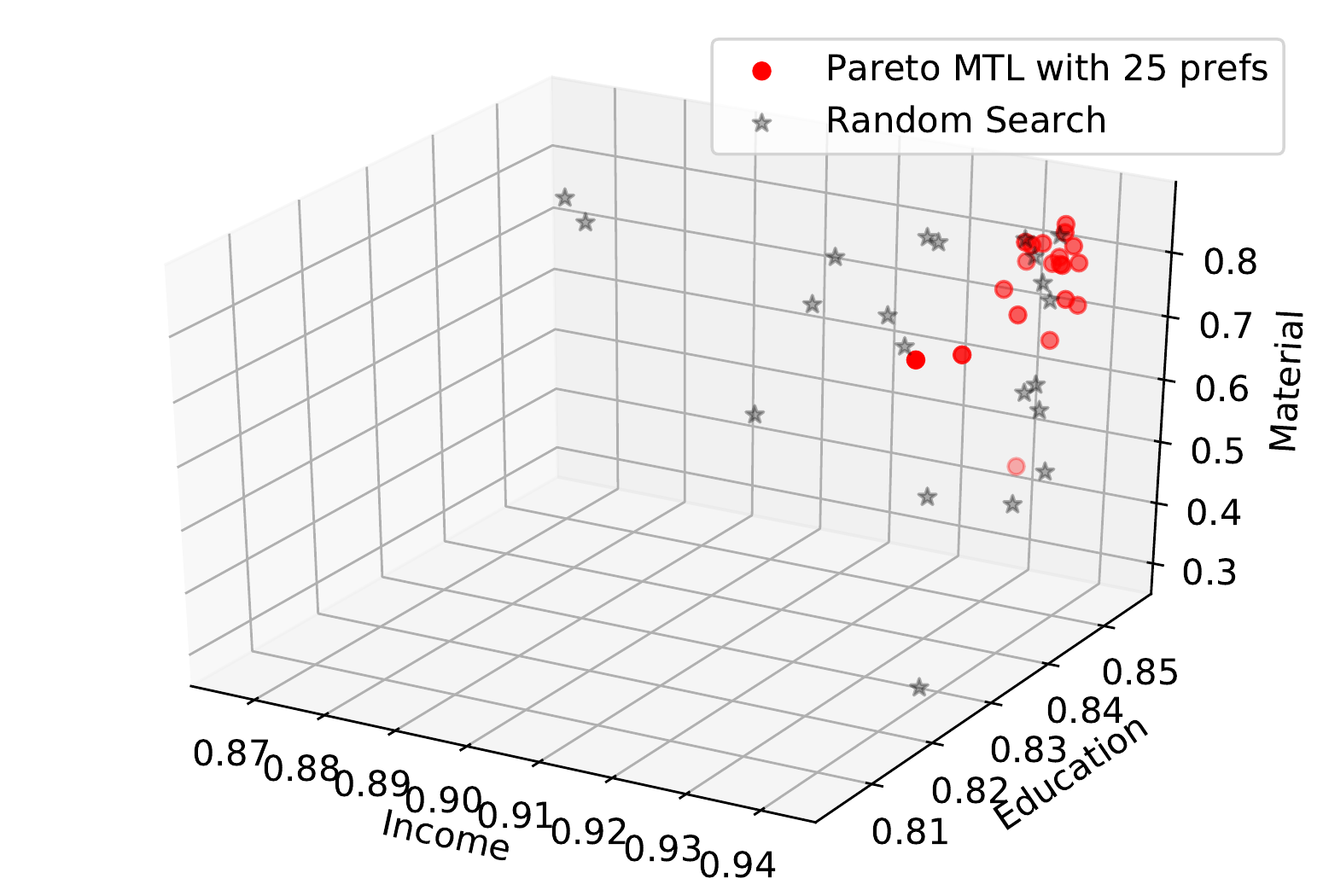}}
\caption{Pareto MTL with different number of preference vectors and linear scalarization with random search on census dataset.}
\label{fig:Census}
\end{figure}

The experiment's results are shown in Fig.~\ref{fig:Census}. We first compare the performance of Pareto MTL with $5$ and $25$ randomly generated preference vectors. From sub-figure (a), it is clear that Pareto MTL with $25$ preference vectors can represent the different trade-offs among three tasks much better. MTL practitioners can easily select their preferred solutions from the obtained results. Pareto MTL with $5$ preference vectors can also provide solutions with different trade-offs, but the only $5$ obtain solutions can not properly represent different optimal trade-offs on the Pareto front. Therefore, a large number of preference vectors (and hence corresponding MTL subproblems) is needed to obtain a set of well-representative trade-offs for a MTL problem with more tasks. We also compare Pareto MTL with the linear scalarization method with random weight research. As shown in sub-figure (b), Pareto MTL's solutions dominate nearly all random search's solutions, which means Pareto MTL has a much better performance on this MTL problem.

In addition to the performance, how to provide information to the practitioner for making decisions is another critical issue for many tasks. Visualizing all solutions with different trade-offs for three tasks is not as clear as for two tasks, and visualization would be much more difficult for more than three tasks. We make a discussion on some potential methods for many tasks in the rest of this section.

\paragraph{Finding representative solutions with preferred trade-offs.} When the preference vectors are fixed and cannot be adaptively adjusted, MTL practitioners can still directly use Pareto MTL to generate different Pareto solutions with their preferred trade-offs for MTL problem with many tasks. As discussed in the experiment section, in Pareto MTL, the subproblem with extreme preference vector (e.g., $(0,1)$ and $(1,0)$) can be explained as auxiliary multi-task learning, where the corresponding task is the preferred main task and the others are auxiliary tasks. Similar explains can also be applied for subproblem with a specific preference vector. In other words, once the MTL practitioners have their preferred trade-off(s) among the tasks, they can directly run Pareto MTL to find diverse Pareto solution(s) corresponding to different preference vectors. If the MTL practitioners do not have any preferred trade-off yet, they can at least run Pareto MTL with a few different preference vectors to obtain a set of diverse Pareto solutions. They can directly choose their preferred solutions or use them to summarize preferred trade-off(s) for another run.

In this section, we run Pareto MTL with only a few preference vectors for solving a MTL with three different tasks. The dataset we use is the UTKFace dataset~\cite{zhifei2017cvpr}, and the MTL problem is to predict human's gender, race, and age based on one image of their faces. We build a deep MTL network with Resnet18 as the encoder and a task-specific fully connected layer for each task.

\begin{table}[H]
    \centering
    \caption{The gender accuracy, race accuracy, and age L1-loss obtained by different algorithms. The best results are highlighted. Pareto MTL can find widely distributed solutions with diverse trade-offs.}
    \label{table:UTK}
    \begin{tabular}{ccccc}
    \hline
    Method      & Reference & Gender            & Race              & Age               \\
                & Vector    & (Accuracy)        & (Accuracy)        & (L1-Loss)         \\ \hline
    Single Task & -         & 0.879157          & 0.781265          & 13.55133          \\ \hline
    Fixed MTL   & $(\frac{1}{3},\frac{1}{3},\frac{1}{3})$   & 0.873982          & 0.783819          & 13.90006          \\ \hline
    GradNorm    & -         & 0.880913          & 0.766011          & 13.79205          \\
    Uncertainty & -         & 0.879657          & 0.770406          & 13.96869          \\
    MOO-MTL     & -         & 0.878013          & 0.778436          & 13.80779          \\ \hline
                & $(\frac{\sqrt{3}}{3},\frac{\sqrt{3}}{3},\frac{\sqrt{3}}{3})$   & 0.878049          & 0.786747          & 13.91812          \\
    Pareto MTL  & $(0,\frac{\sqrt{2}}{2},\frac{\sqrt{2}}{2})$   & \textbf{0.885552} & 0.754095          & 14.41414          \\
                & $(\frac{\sqrt{2}}{2},0,\frac{\sqrt{2}}{2})$   & 0.872242          & \textbf{0.792814} & 14.32156          \\
                & $(\frac{\sqrt{2}}{2},\frac{\sqrt{2}}{2},0)$   & 0.866895          & 0.762251          & \textbf{13.51771} \\ \hline
    \end{tabular}
\end{table}

\begin{figure}[H]
\centering
\subfloat[gender v.s. race]{\includegraphics[width = 0.33\textwidth]{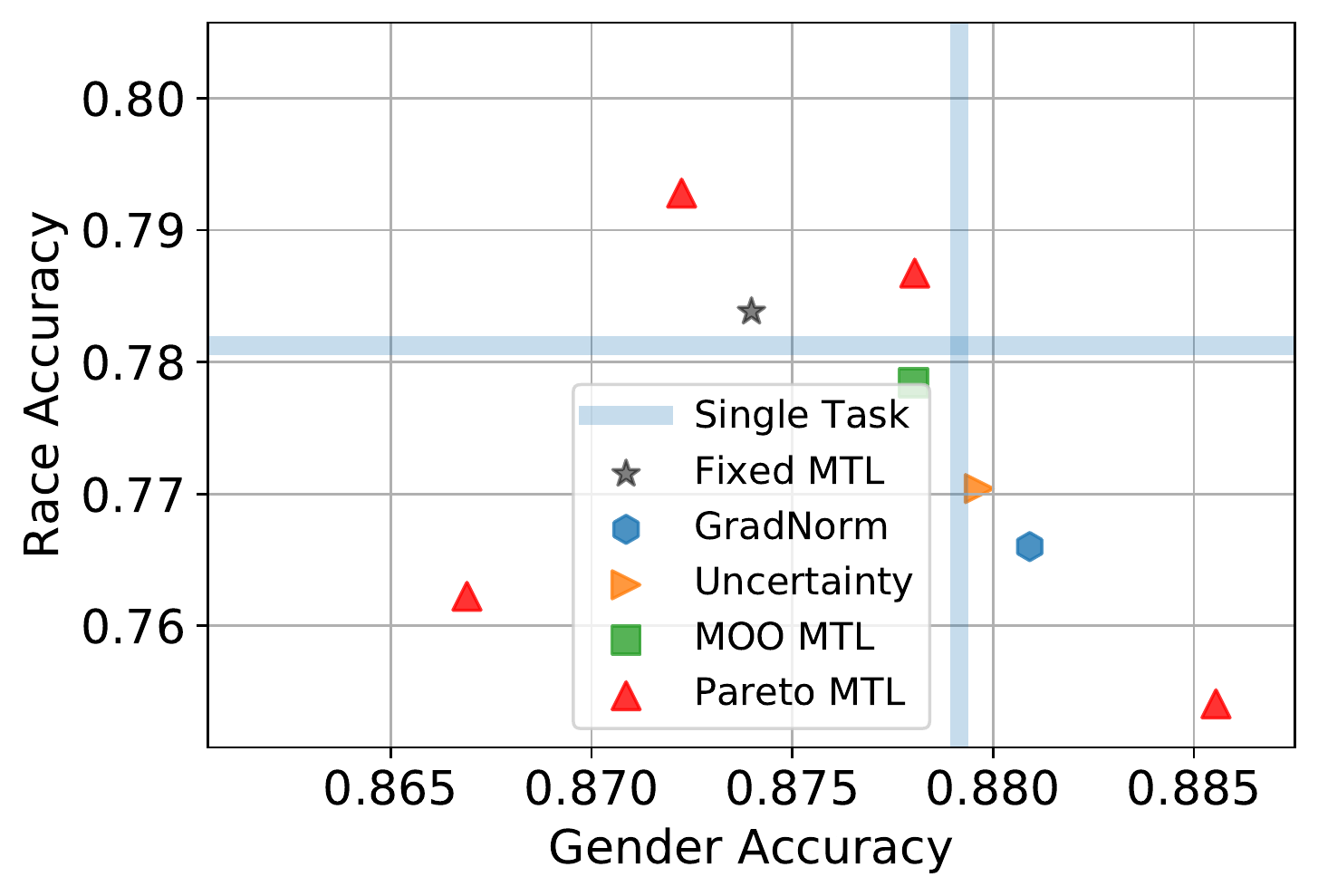}}
\subfloat[gender v.s. age]{\includegraphics[width = 0.33\textwidth]{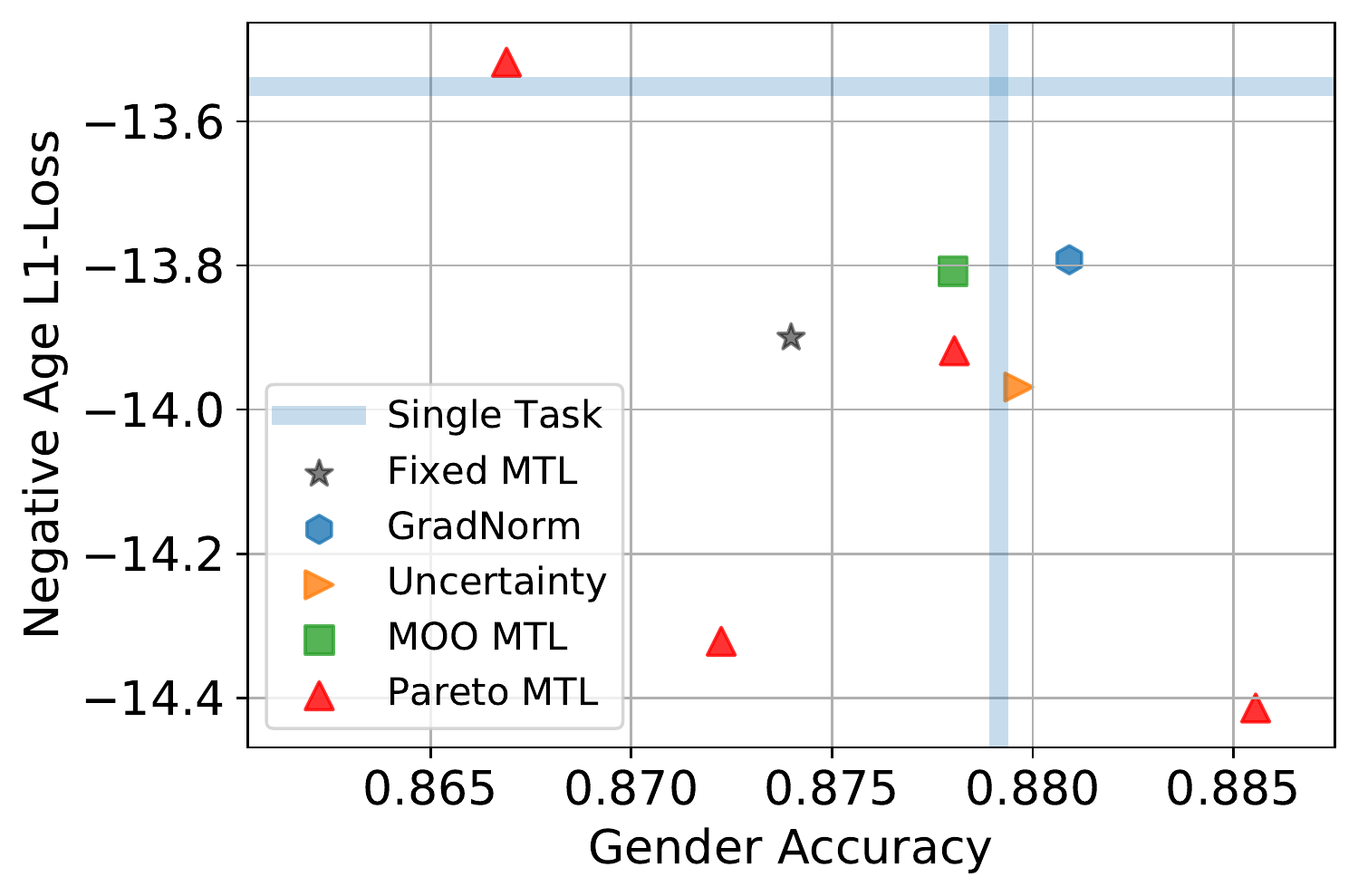}}
\subfloat[race v.s. age]{\includegraphics[width = 0.33\textwidth]{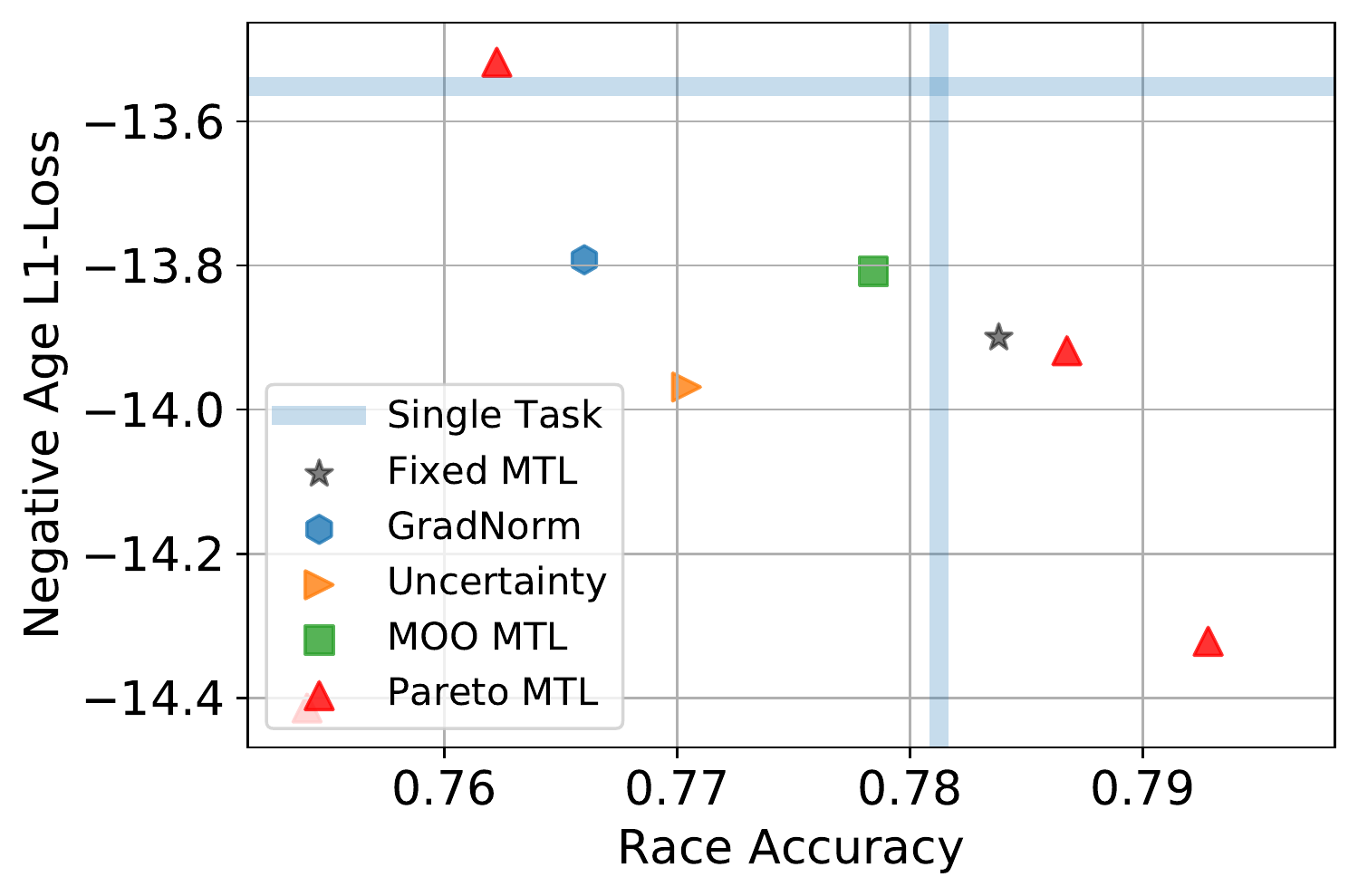}} \\
\caption{The 2-D projections for the results obtained by different algorithms. We report the negative age L1-loss for the sake of consistency. Pareto MTL can provide solutions with diverse trade-offs.}

\label{fig:UTK}
\end{figure}

The experiment result is shown in Table.\ref{table:UTK} and Fig.\ref{fig:UTK}. It confirms that Pareto MTL with a few specific preference vectors can still find representative Pareto solutions for the practitioner's preferred trade-offs.

\textbf{Learning to find preferred Pareto solution(s).} One advantage of Pareto MTL over MOO-MTL is its ability to incorporate preference information even with only a single run (solving one subproblem, but still need a set of preference vectors to divide the objective space). Recently, some learning-based methods have been proposed to solve MTL problems~\cite{zhang2018learning,guo2018dynamic}. It is possible to propose learning-based Pareto MTL for dynamically adjusting the preference vectors to incorporate the MTL practitioner's preference, and to guide the solutions search to their preferred smaller subspace for a MTL problem.

\textbf{Methods from the multi-objective optimization community.} By formulating the MTL with many tasks as a multi-objective optimization problem, we get a many-objective optimization problem, which is indeed a popular research topic in the multi-objective optimization community~\cite{fleming2005many,ishibuchi2008evolutionary,hadka2013borg,deb2013evolutionary}. How to adopt the techniques proposed from the multi-objective optimization community to solve the MTL problem with many tasks is a potential research direction.

\clearpage

\section{The Adaptive Weight Vectors}

\begin{figure*}[h]
\centering
\subfloat[ParetoMTL (1,0) ]{\includegraphics[width = 0.33\textwidth]{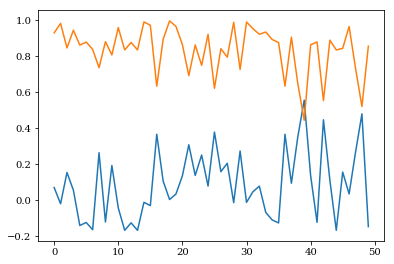}}
\subfloat[ParetoMTL $(\sqrt{2}/2,\sqrt{2}/2)$]{\includegraphics[width = 0.33\textwidth]{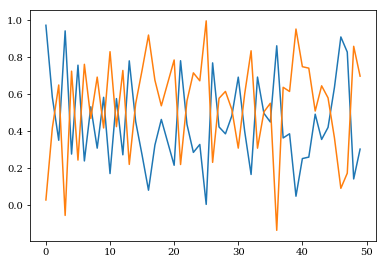}}
\subfloat[ParetoMTL (0,1)]{\includegraphics[width = 0.33\textwidth]{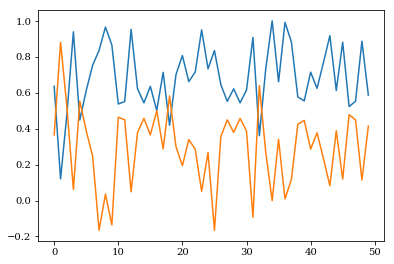}} \\
\subfloat[MOO-MTL]{\includegraphics[width = 0.33\textwidth]{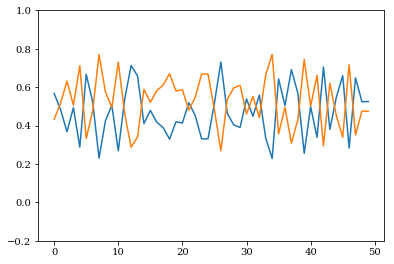}}
\subfloat[GradNorm]{\includegraphics[width = 0.33\textwidth]{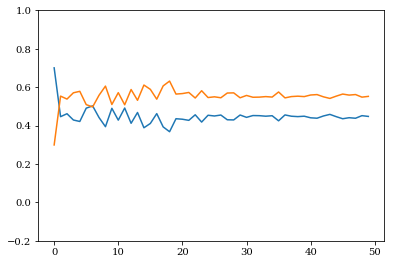}}
\subfloat[Fixed Weight]{\includegraphics[width = 0.33\textwidth]{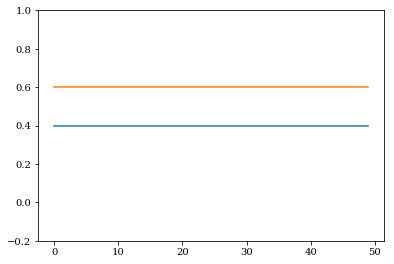}}
\caption{\textbf{The adaptive weight vectors for different algorithms during the training process for MultiMNIST experiment.} Pareto MTL behaves differently with different reference vectors. The other algorithms with adaptive weights assignment try to balance different tasks.}
\label{weight_trend}
\end{figure*}

In the paper, we show that MOO-MTL and the proposed Pareto MTL algorithm can be reformulated as a linear scalarization of different tasks with adaptive weights assignment where the Pareto MTL algorithm can be rewritten as:
\begin{eqnarray}
    \label{update_mtl_c_app}
    \mathcal{L}(\theta_t) = \sum_{i=1}^{m} \alpha_i  \mathcal{L}_i(\theta_t), ~\text{where }  \alpha_i = \lambda_i + \sum_{j \in I_{\epsilon(\theta)}} \beta_j (\vu_{ji} - \vu_{ki}),
\end{eqnarray}

In this section, we compare the adaptive weight vectors for different algorithms during the training process. As shown in Fig.~\ref{weight_trend}, Pareto MTL has clearly different weight adaption strategies for subproblems with different preference vectors, while MOO-MTL and GradNorm always try to balance different tasks.

From the view point of linear scalarization with adaptive weights, the surrogate loss for MOO-MTL, GradNorm and Uncertainty can be written as $\mathcal{L}(\theta_t) = \sum_{i=1}^{m} \lambda_i  \mathcal{L}_i(\theta_t)$. Different methods have their own strategy to adapt the weight $\lambda_i$ to balance the loss function $\mathcal{L}_i(\theta_t)$. For Pareto MTL, the weight vector now has the form $\alpha_i = \lambda_i + \sum_{j \in I_{\epsilon(\theta)}} \beta_j (\vu_{ji} - \vu_{ki})$. While the parameter $\lambda_i$ is still for balancing different tasks, the preference term $\sum_{j \in I_{\epsilon(\theta)}} \beta_j (\vu_{ji} - \vu_{ki})$ will guide the Pareto MTL to its corresponding preference vector. As shown in Fig.~\ref{weight_trend}, Pareto MTL will bias the search to a specific task when it has extreme preference vectors (e.g., $(0,1)$ and $(1,0)$), and it will try to balance different tasks with a balance preference vector (e.g., $(\sqrt{2}/2,\sqrt{2}/2)$).

When all constraints are inactivated (e.g., $I_{\epsilon(\theta)} = \emptyset $), Pareto MTL has a feasible solution right in the assigned sub-region for a given subproblem. In this case, the surrogate loss could be reduced to $\mathcal{L}_i(\theta_t) = \sum_{i=1}^{m} \lambda_i  \mathcal{L}_i(\theta_t)$ which is the same as MOO-MTL, and Pareto MTL will try to find a balanced solution in the assigned sub-region.

Pareto MTL is not mutually exclusive with other adaptive weight strategies such as GradNorm~\cite{chen2018grad} and Uncertainty~\cite{kendall2017multi}, especially when it can be reformulated as the linear scalarization method. For MTL problem with highly unbalanced tasks with different difficulties, it is possible to first balance all tasks with some adaptive weight strategies, and then use Pareto MTL to find a set of Pareto solutions for the balanced tasks. We will discuss the Pareto MTL's performance on tasks with different difficulties in the next section.

\section{Pareto MTL with Different Tasks Difficulties}

\begin{figure*}[h]
\centering
\subfloat[Linear ($a_1 = 2, a_2 =1$)]{\includegraphics[width = 0.33\textwidth]{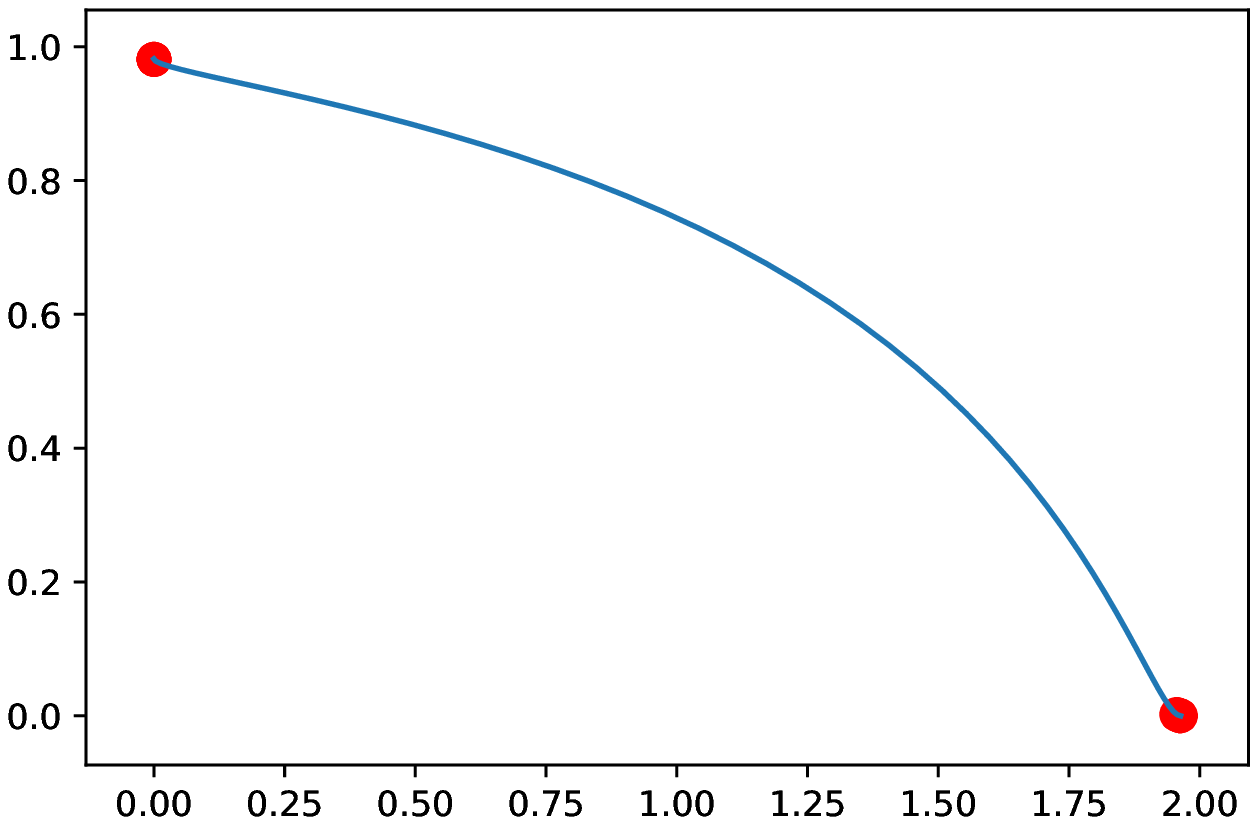}}
\subfloat[MOO MTL ($a_1 = 2, a_2 = 1$)]{\includegraphics[width = 0.33\textwidth]{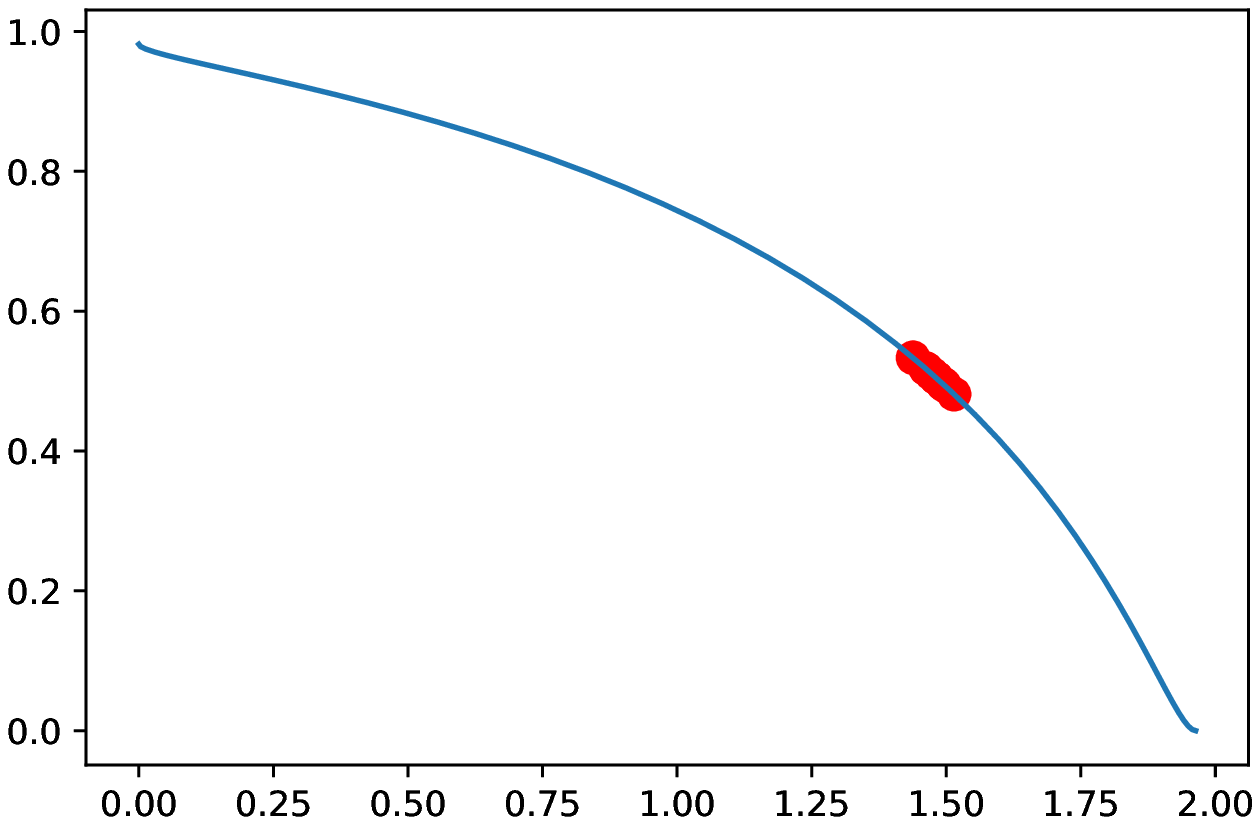}}
\subfloat[Pareto MTL ($a_1 = 2, a_2 = 1$)]{\includegraphics[width = 0.33\textwidth]{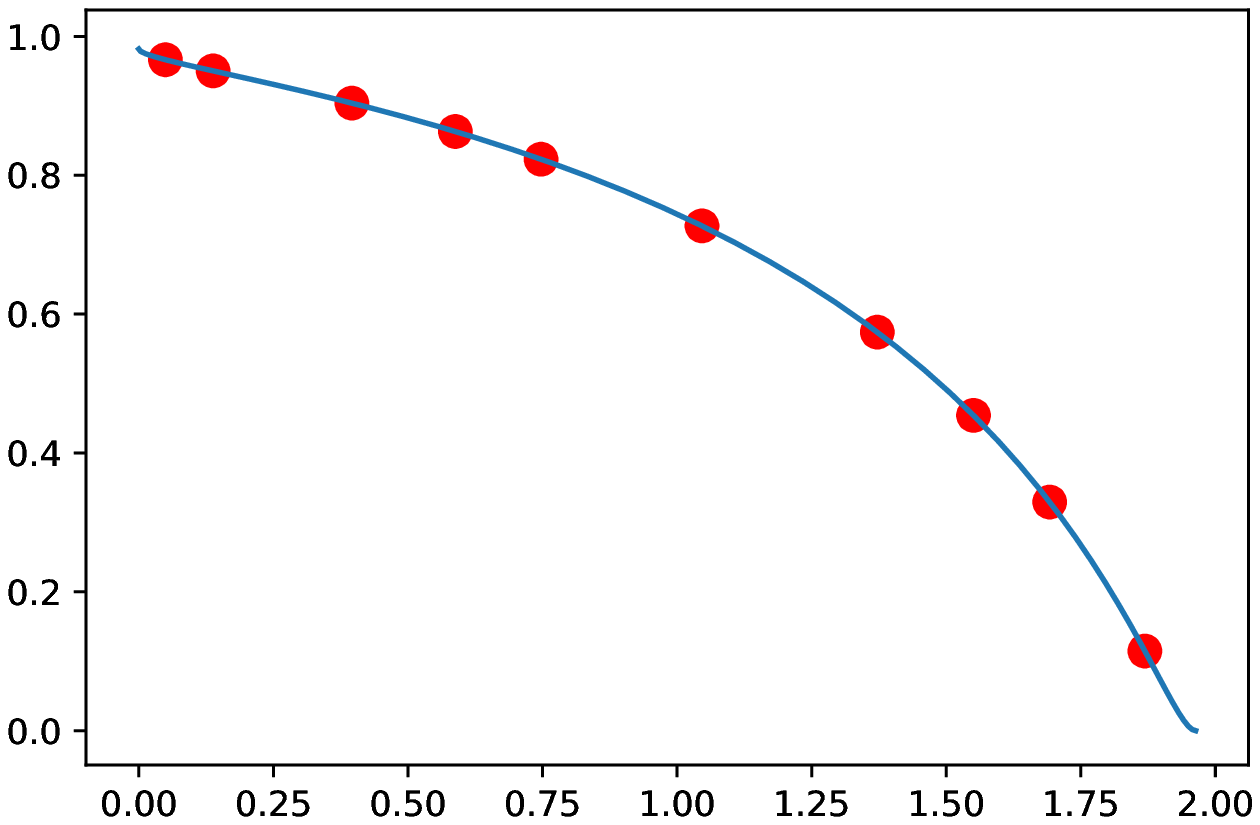}}
\\
\subfloat[Linear ($a_1 = 10, a_2 =1$)]{\includegraphics[width = 0.33\textwidth]{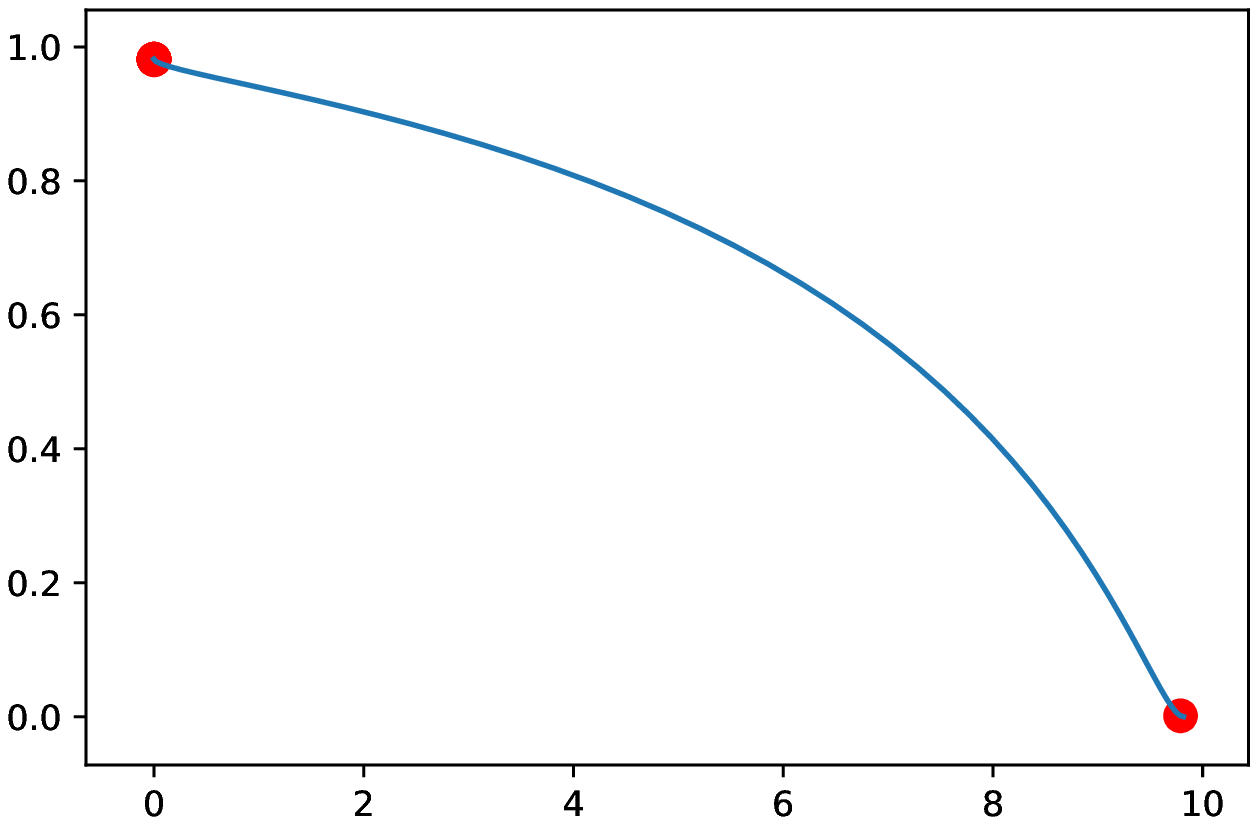}}
\subfloat[MOO MTL ($a_1 = 10, a_2 =1$)]{\includegraphics[width = 0.33\textwidth]{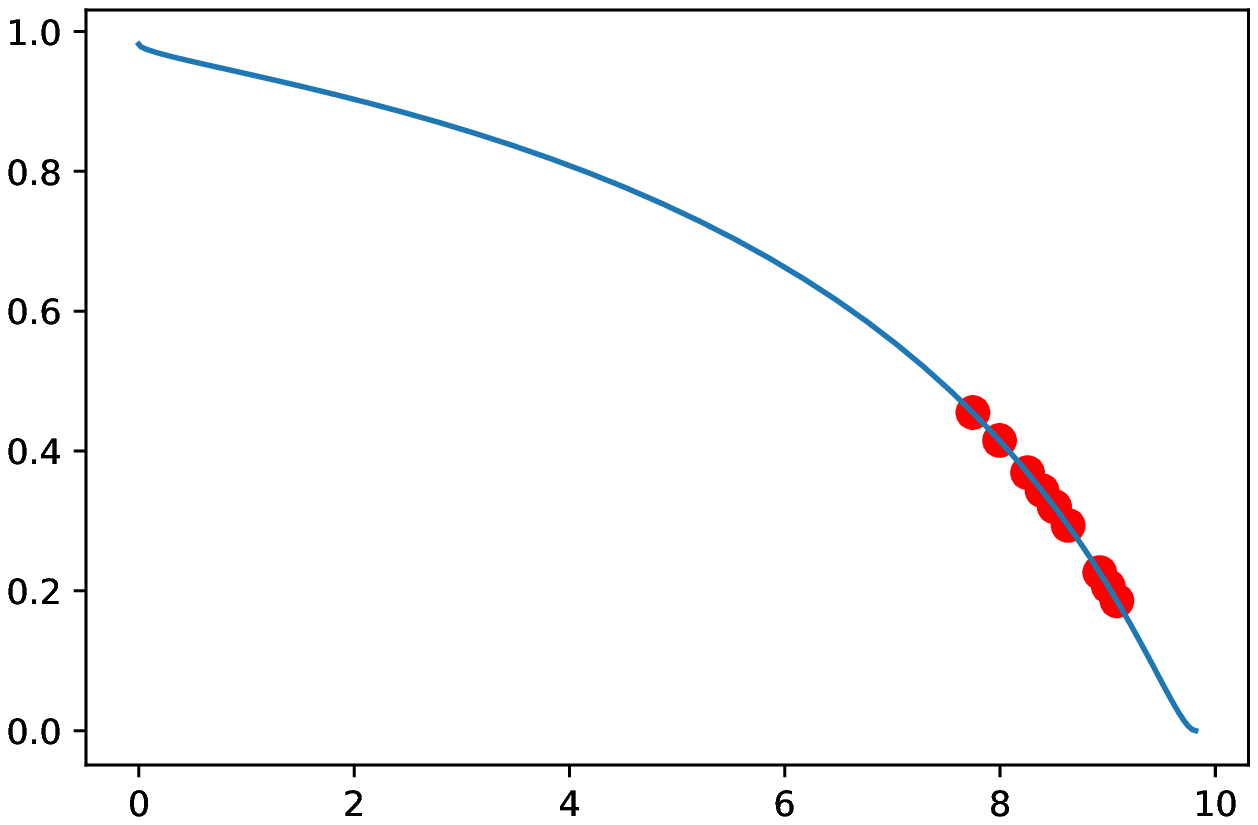}}
\subfloat[Pareto MTL ($a_1 = 10, a_2 =1$)]{\includegraphics[width = 0.33\textwidth]{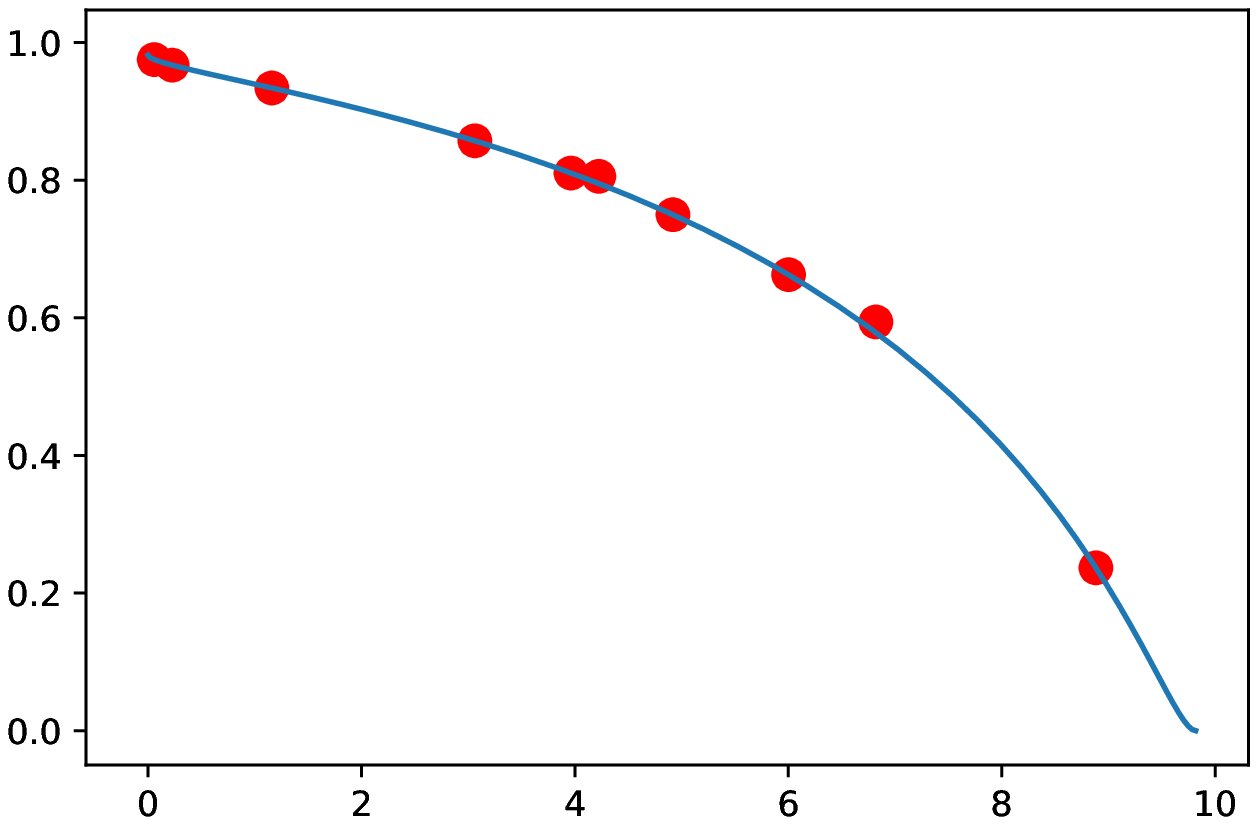}}
\\
\subfloat[Linear ($a_1 = 50, a_2 =1$)]{\includegraphics[width = 0.33\textwidth]{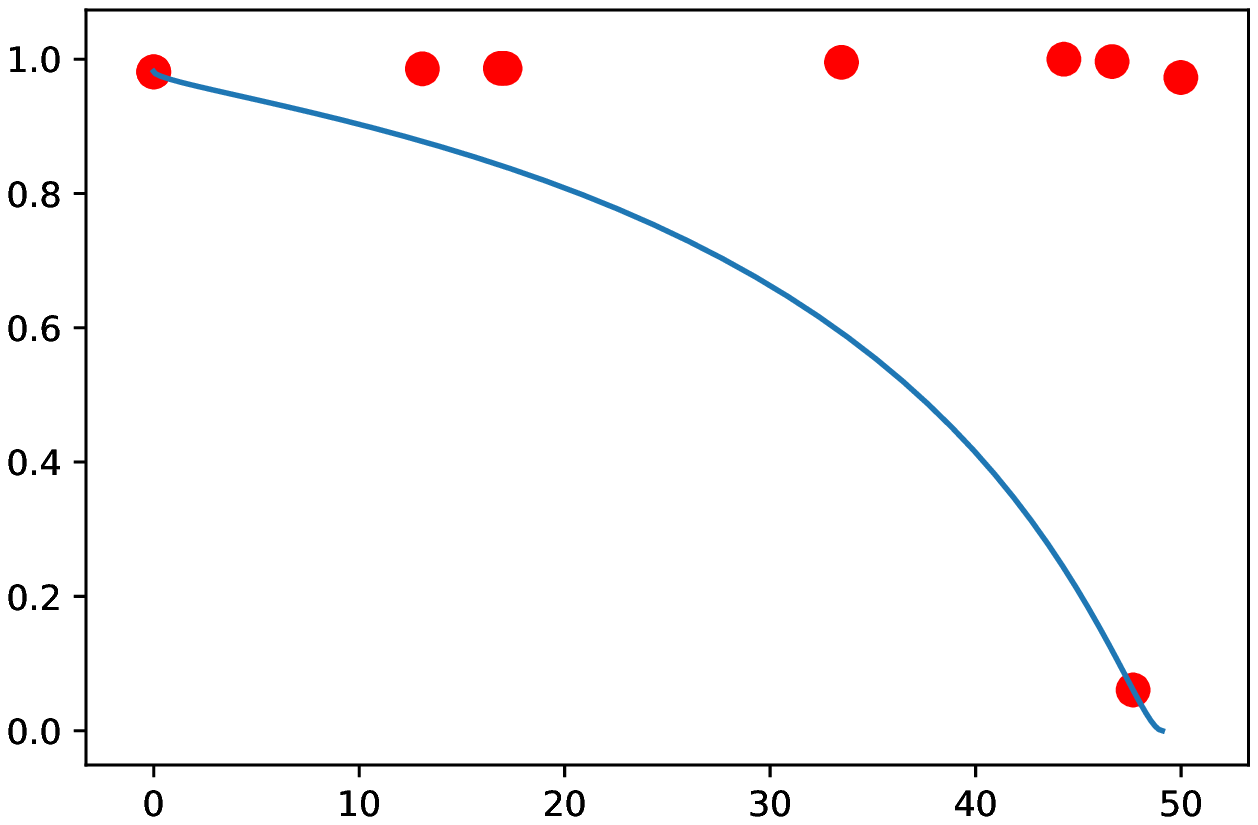}}
\subfloat[MOO MTL ($a_1 = 50, a_2 =1$)]{\includegraphics[width = 0.33\textwidth]{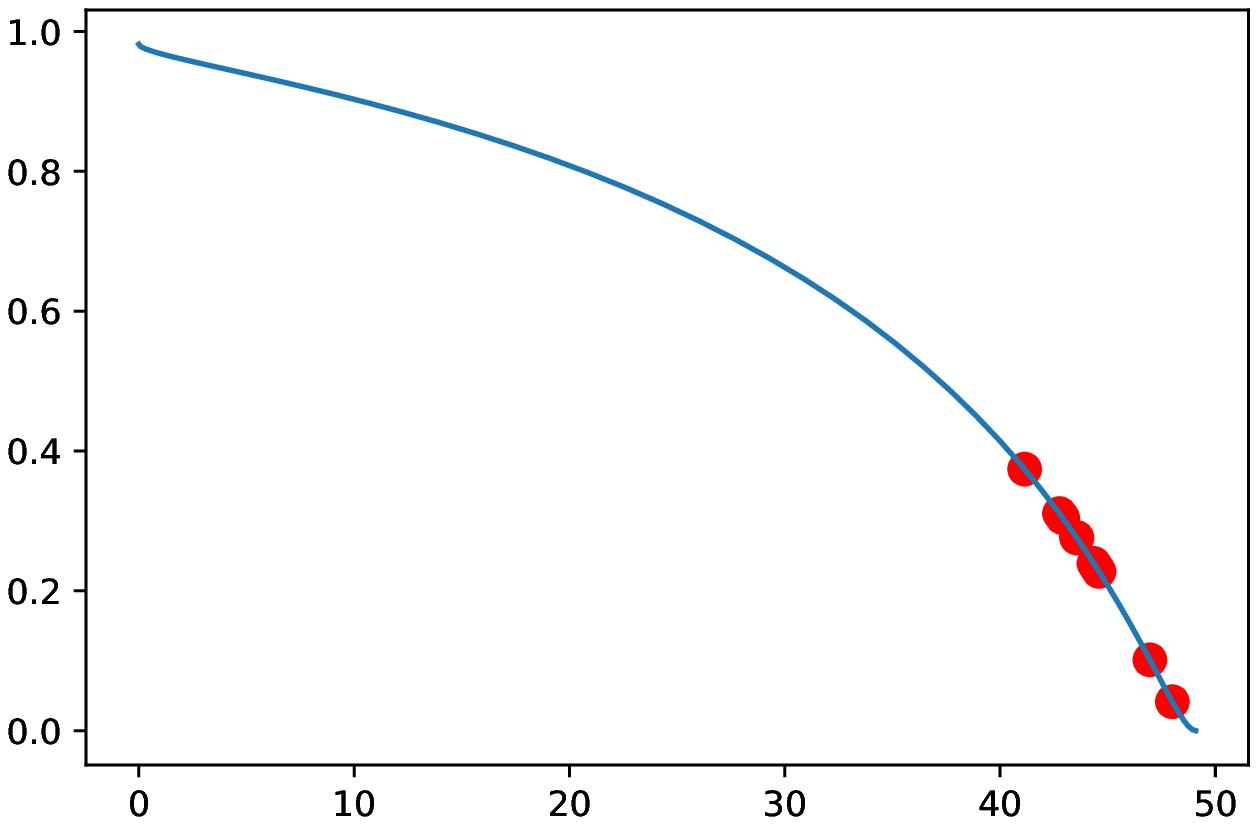}}
\subfloat[Pareto MTL ($a_1 = 50, a_2 =1$)]{\includegraphics[width = 0.33\textwidth]{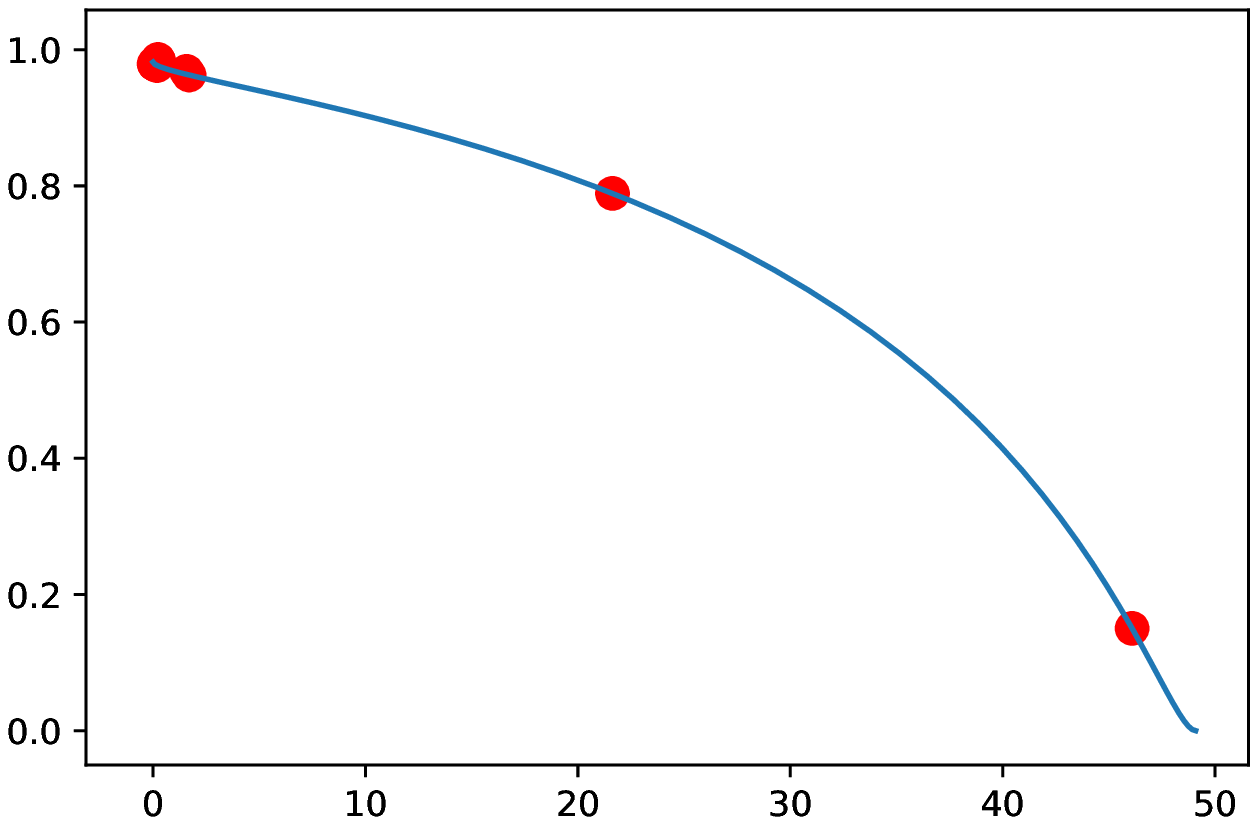}}

\caption{The convergence behaviours of different algorithms on a synthetic example with different tasks difficulties. Pareto MTL can find a set of widely distributed Pareto solutions on problems with low or medium difficulty unbalance level (($a_1 = 2, a_2 =1$) and ($a_1 = 10, a_2 =1$)), but its performance gets worse for problem with high difficulty unbalance level ($a_1 = 50, a_2 =1$).}
\label{PF_toy_example_app}
\end{figure*}

Pareto MTL implicitly assumes the tasks in a MTL problem should have similar difficulties, and uses the set of widely distributed unit vectors as the preference vectors. However, its performance might be deteriorated for tasks with extremely different difficulties. It is hard to control the tasks' difficulties in real-world MTL applications manually. To clearly demonstrate the performance of Pareto MTL, we test it on the following synthetic example:
\begin{eqnarray}
    \label{toy_example_app}
    \begin{aligned}
        &\min f_1(\vx) = \alpha_1  - \alpha_1 \exp{(- \sum_{i=1}^{d} (x_d - \frac{1}{\sqrt{d}})^2)}, \\
        &\min f_2(\vx) = \alpha_2 - \alpha_2 \exp{(- \sum_{i=1}^{d} (x_d + \frac{1}{\sqrt{d}})^2)},
    \end{aligned}
\end{eqnarray}
where $f_1(\vx)$ and $f_2(\vx)$ are two objective functions to be minimized at the same time and $\vx = (\vx_1, \vx_2,..., \vx_d)$ is the $d$ dimensional decision variable. This problem has a concave Pareto front on the objective space. The two objective function can have different difficulty levels controlled by the parameters $\alpha_i$. If $\alpha_1 = \alpha_2$, the two tasks have similar difficult level, which is the synthetic example we have in the main paper. If $\alpha_1 > \alpha_2$, the task one could be "easier" since it has a larger gradient value, and task 2 could be "easier" if $\alpha_1 < \alpha_2$. We notice the difficulty measurement for real-world multi-task learning problem would be much more complicated. Here we focus on the much simplified version and left the analysis for real-world problems in the future.

\textbf{Performance for Problems with Unbalanced Difficulties.} The results on problems with different levels of unbalanced difficulties are shown in Fig.~\ref{PF_toy_example}. As in the balanced difficult case, the linear scalarization and MOO-MTL approach can not obtain a set of well-representative solutions for all problems. Pareto MTL can find a set of widely distributed Pareto solutions on problems with low or medium difficulty unbalance level (($a_1 = 2, a_2 =1$) and ($a_1 = 10, a_2 =1$)), but its performance gets worse for problem with high difficulty unbalance level ($a_1 = 50, a_2 =1$).

\textbf{Different Biases for MOO-MTL and Pareto MTL.} Another interesting observation in this experiment is that MOO-MTL and Pareto MTL will be biased to different tasks in the same unbalance problem. For example, in the highly unbalanced problem ($a_1 = 50, a_2 =1$), MOO-MTL is biased to the "easier" Task 1 since it has a much larger absolute gradient value. Most solutions found by Pareto MTL, however, are biased to the "harder" Task 2. When the tasks have different difficulty levels, the decomposed sub-region would be highly unbalanced for evenly distributed preference vectors, and the solutions would be attracted by a few preference vector much easier. In this case, most solutions are attracted by the preference vector corresponding to task 2 rather than task 1.

As discussed in the previous section, combining Pareto MTL and other adaptive weight methods to balance different tasks would be one possible method for tackling different levels of task difficulties. Learning-based self-adaptive method methods would be another important research direction.

\section{Preference Vector Assignment}
\begin{figure*}[h]
\centering
\subfloat[]{\includegraphics[width = 0.33\textwidth]{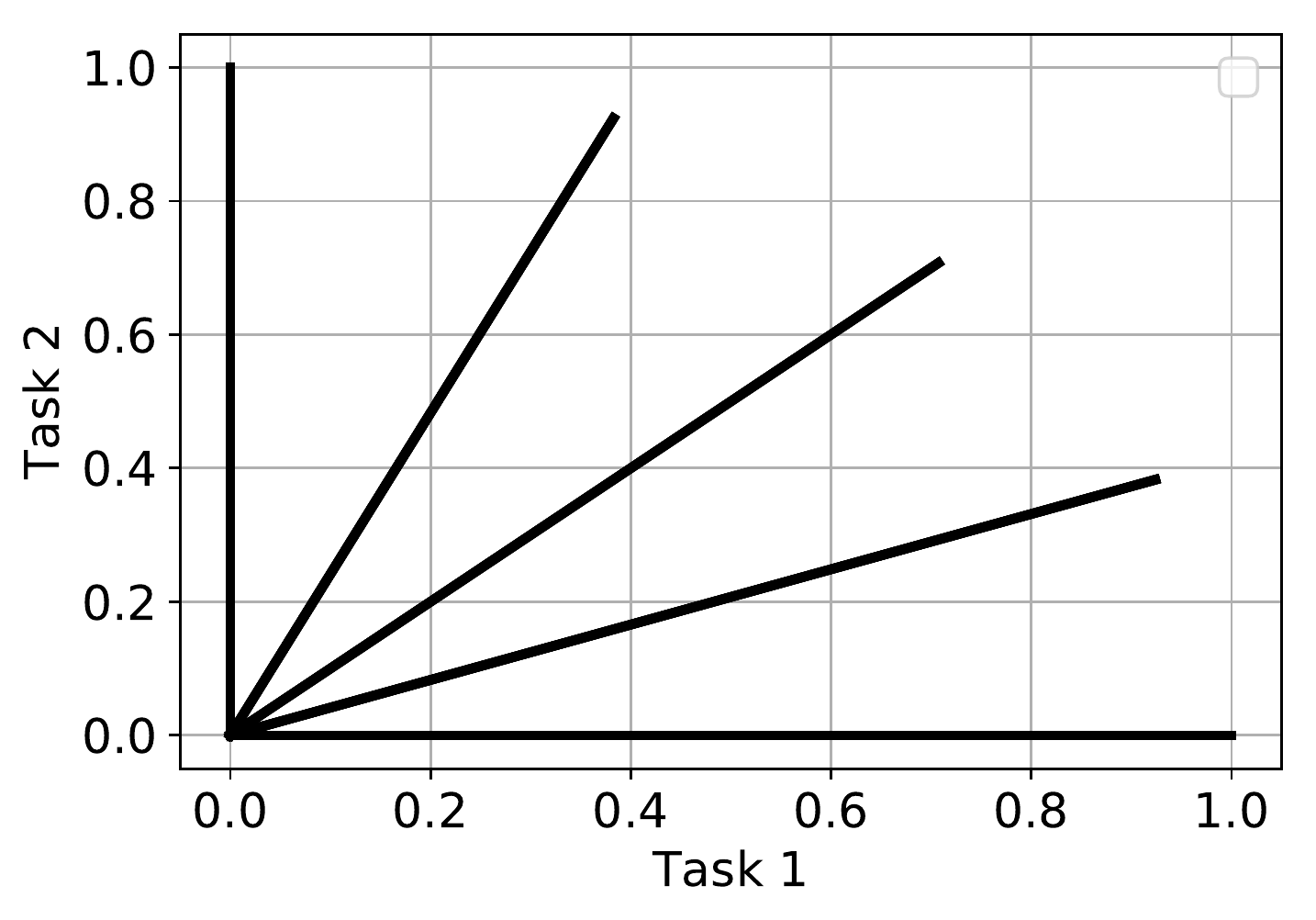}}
\subfloat[]{\includegraphics[width = 0.33\textwidth]{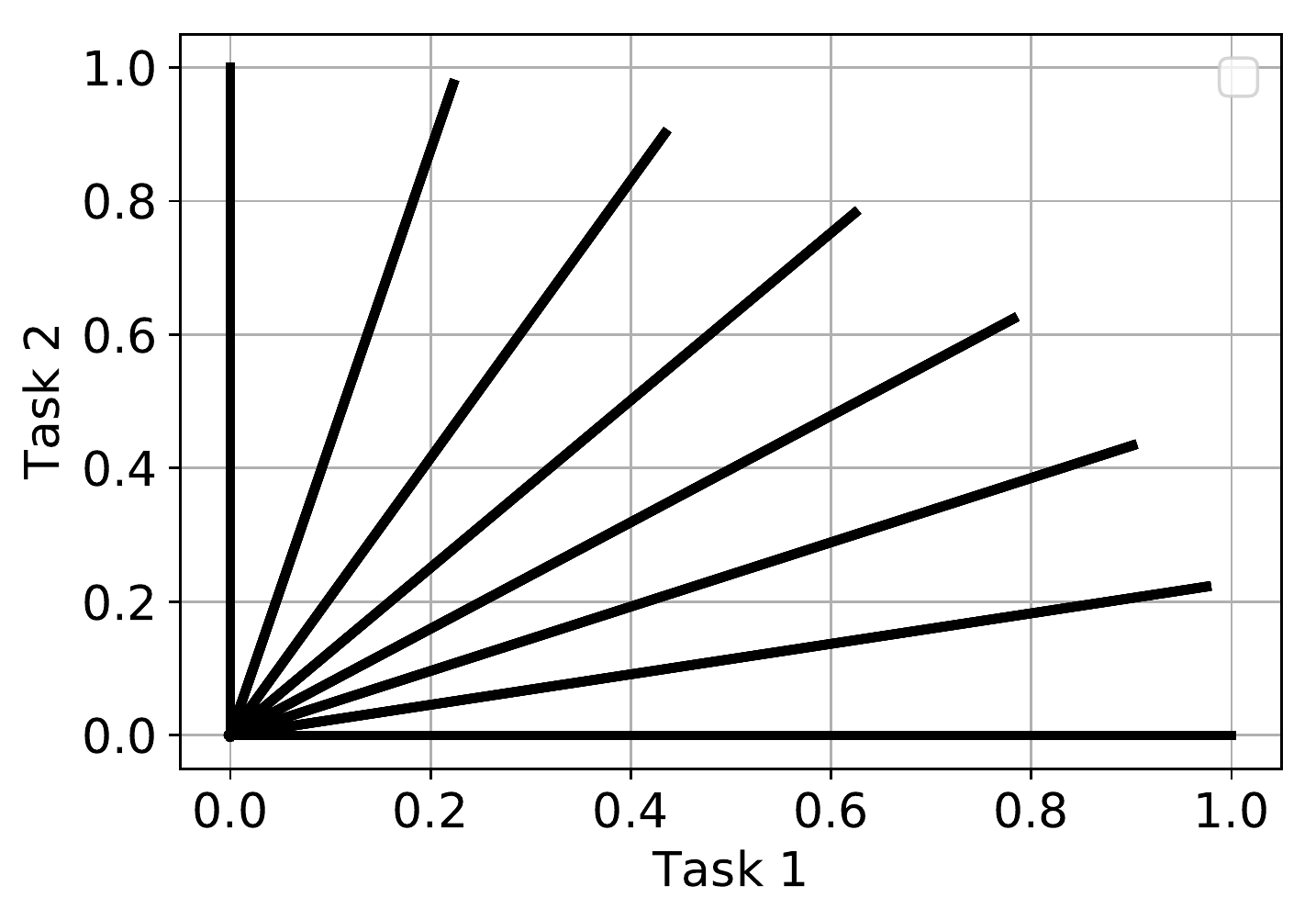}}
\subfloat[]{\includegraphics[width = 0.33\textwidth]{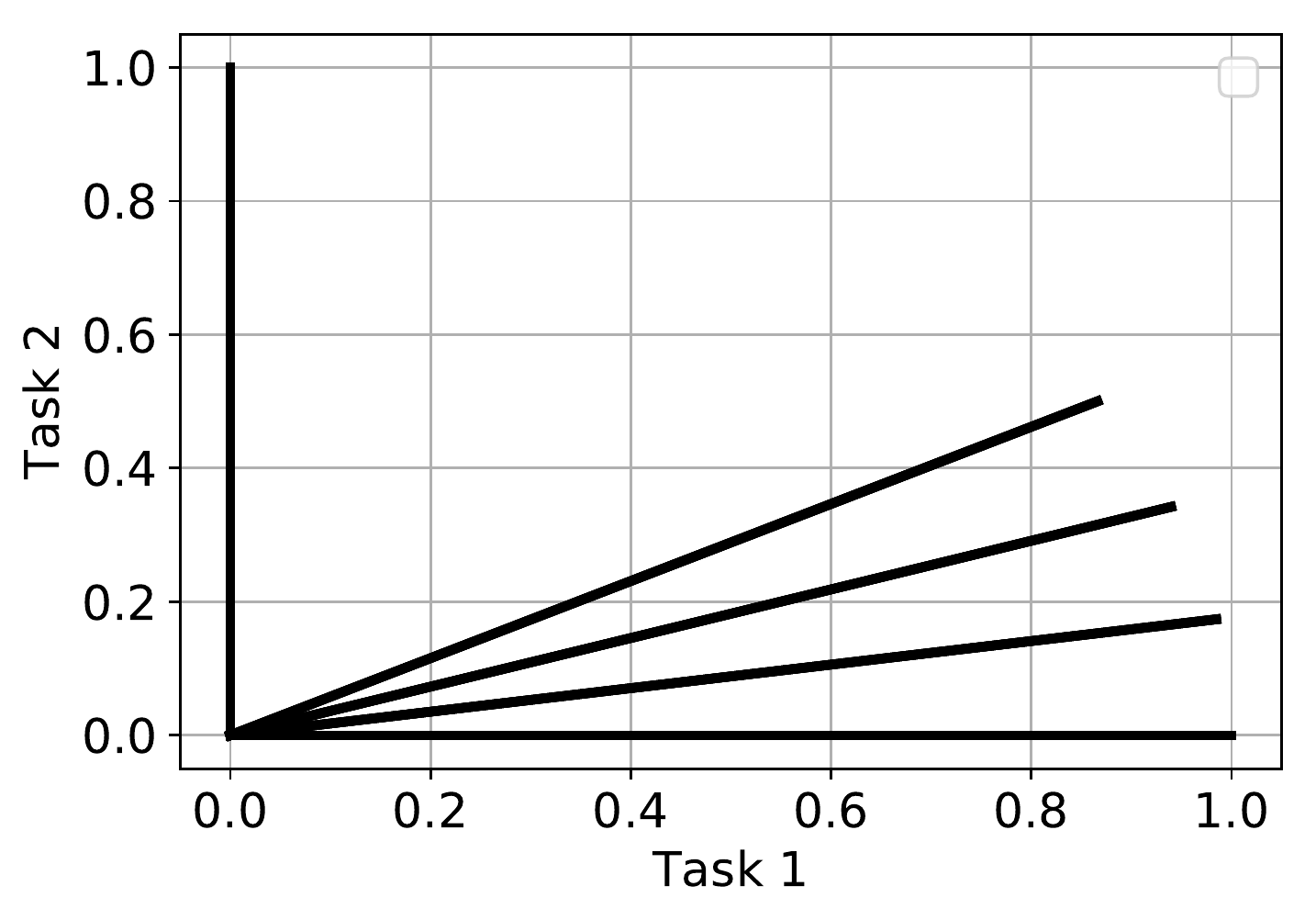}}
\caption{\textbf{Three different sets of preference vector:} (a) 5 evenly distributed unit preference vectors; (b) 8 evenly distributed preference unit vectors; and (c) 5 biased unit preference vectors. }
\label{preference_vector}
\end{figure*}

The final distribution of the solutions obtained by Pareto MTL depends on both the preference vectors and the shape of the Pareto front. Even for tasks with similar difficulties, it is still important to properly assign the set of preference vectors for Pareto MTL. When we do not have any prior information for a given MTL problem, it is reasonable to decompose the objective space for different tasks with a set of evenly distributed preference vectors as shown in Fig.~\ref{preference_vector} (a)(b). When the MTL practitioners have their own preference for a given MTL problem, they can feel free to use a set of biased preference vectors as in Fig.~\ref{preference_vector} (c).

However, it is hard to tell whether the assigned preference vectors would be the optimal one before the actual run of Pareto MTL. We run Pareto MLT multiple times with a different set of randomly generated unit preference vectors, and show the results in Fig.~\ref{MOPMTL}. Pareto MTL with different sets of random preference vectors can consistently generate well-distributed solutions, but the accuracy performance would be better or worse than the default uniform setting. Getting stuck in bad local Pareto optima would be one possible reason for some inferior performance, since Pareto MTL can only guarantee locally restricted Pareto optimality. Our preliminary experiment results also show that too close preference vectors (and hence narrow region $\Omega_k$) could also lead to worse performance.

\begin{figure}[h]
    \centering
    \includegraphics[width= 0.50 \linewidth]{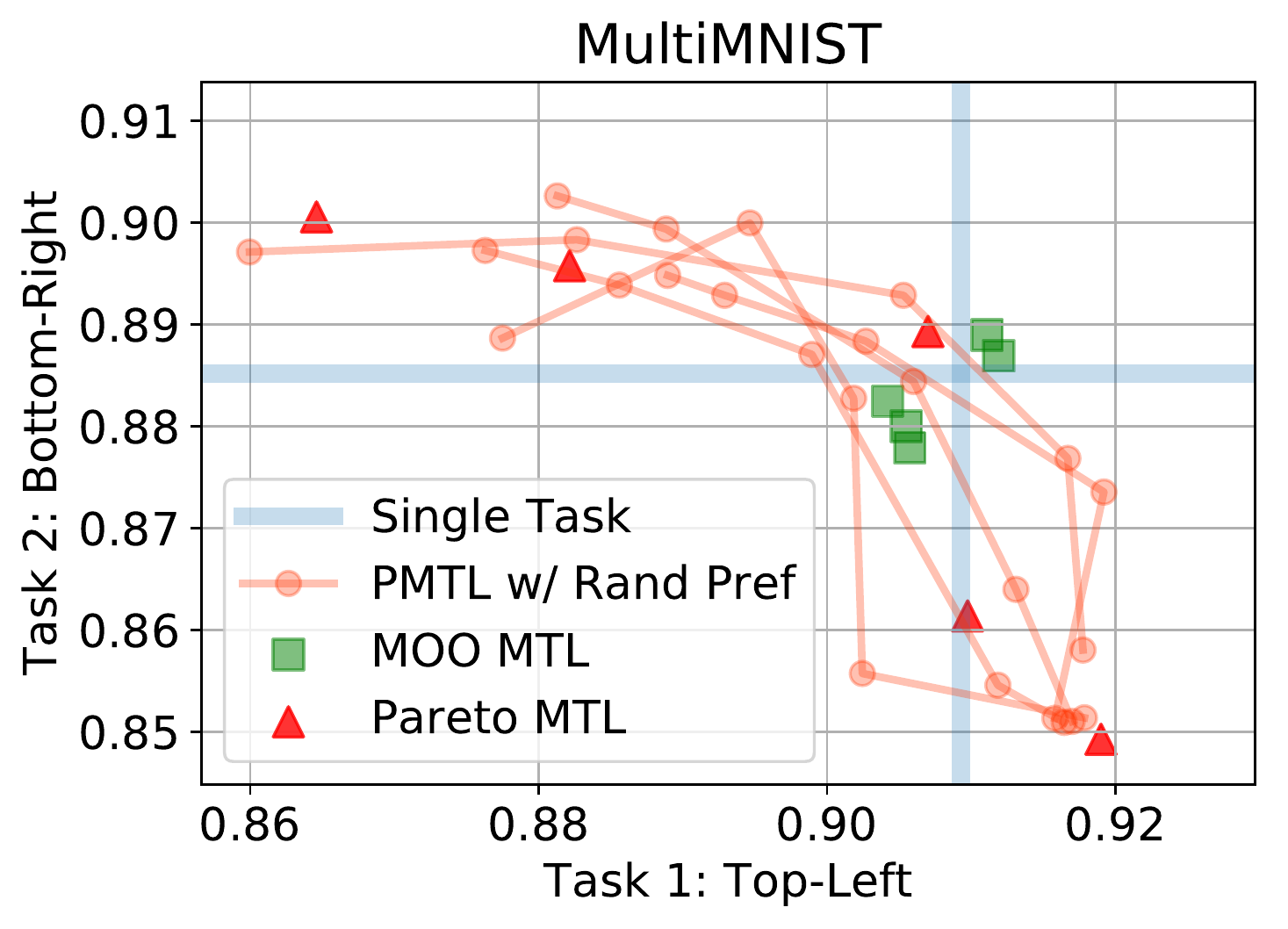}
    \caption{Pareto MTL with different sets of randomly generated unit preference vectors.}
    \label{MOPMTL_app}
\end{figure}

How to efficiently set the preference vectors based on the user's preference or any prior information could be an interesting topic. A strategy to adaptively set or change the preference vectors during the multi-task learning process to incorporate the practitioner's preference or better explore the objective space is another possible extension.

\section{The Gap between Optimization and Generalization}

\begin{figure}[H]
\centering
\subfloat[MultiMNIST: Train Loss]{\includegraphics[width = 0.33\textwidth]{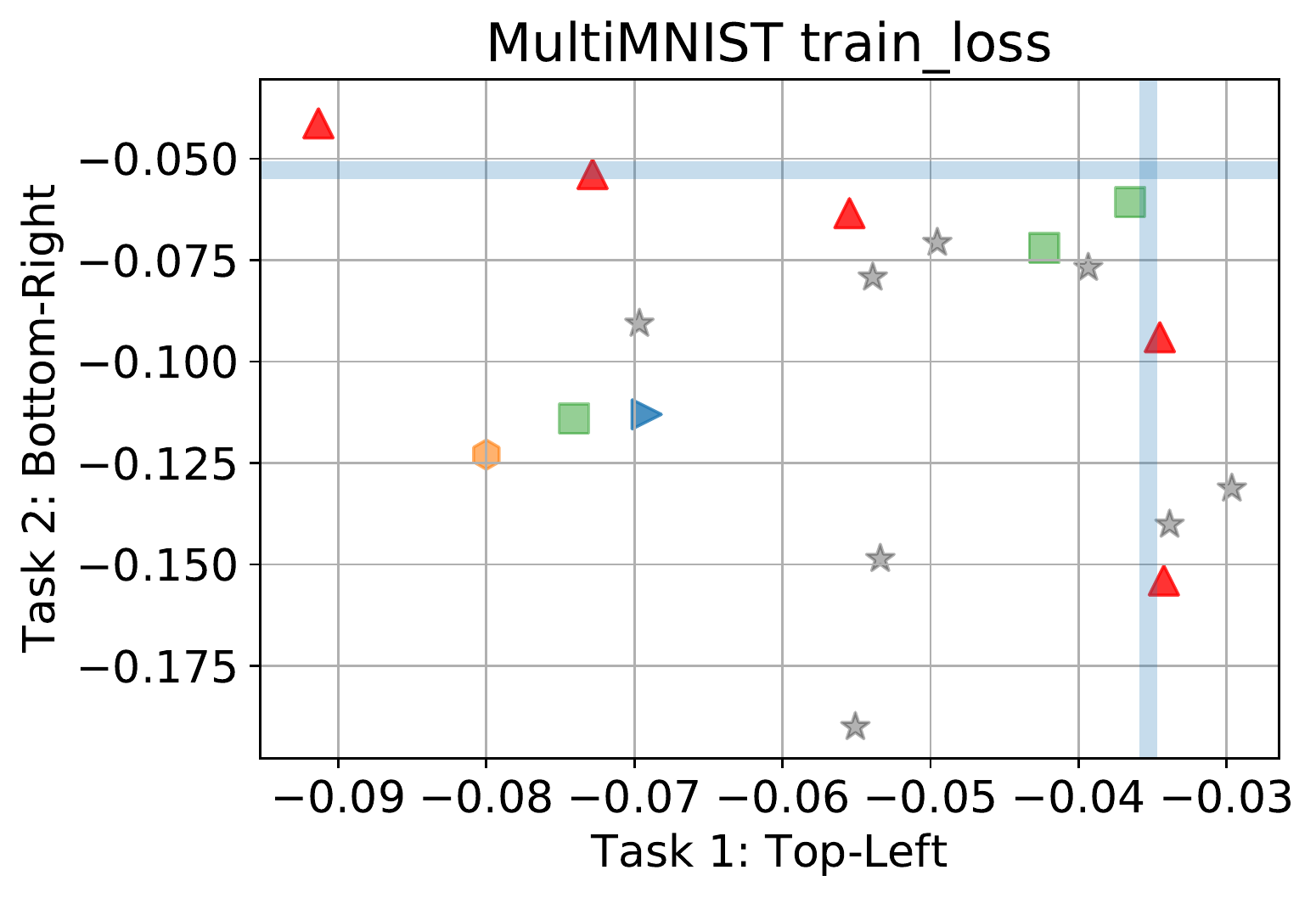}}
\subfloat[MultiMNIST: Train Accuracy]{\includegraphics[width = 0.33\textwidth]{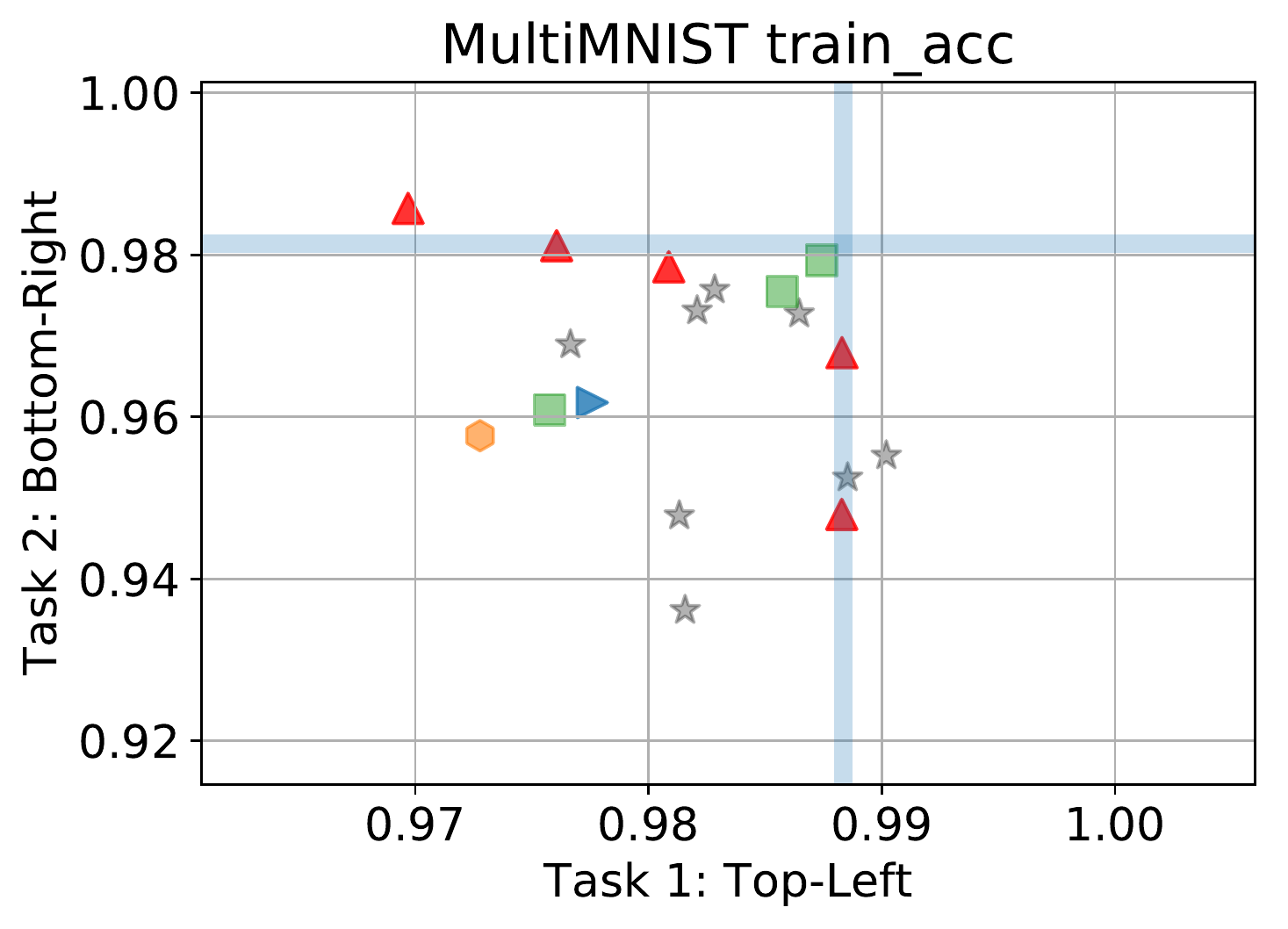}}
\subfloat[MultiMNIST: Test Accuracy]{\includegraphics[width = 0.33\textwidth]{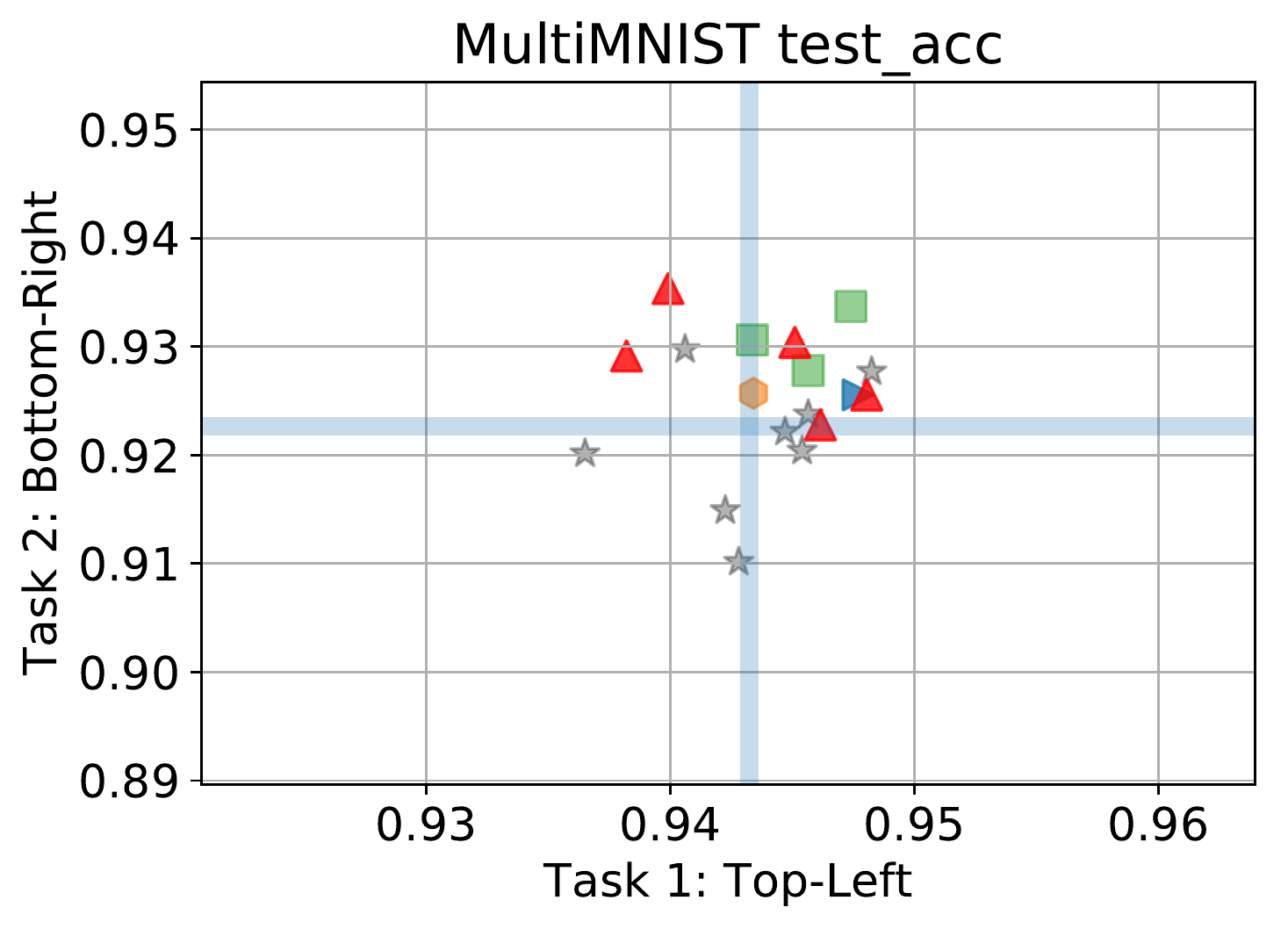}} \\
\subfloat[MultiFashion: Train Loss]{\includegraphics[width = 0.33\textwidth]{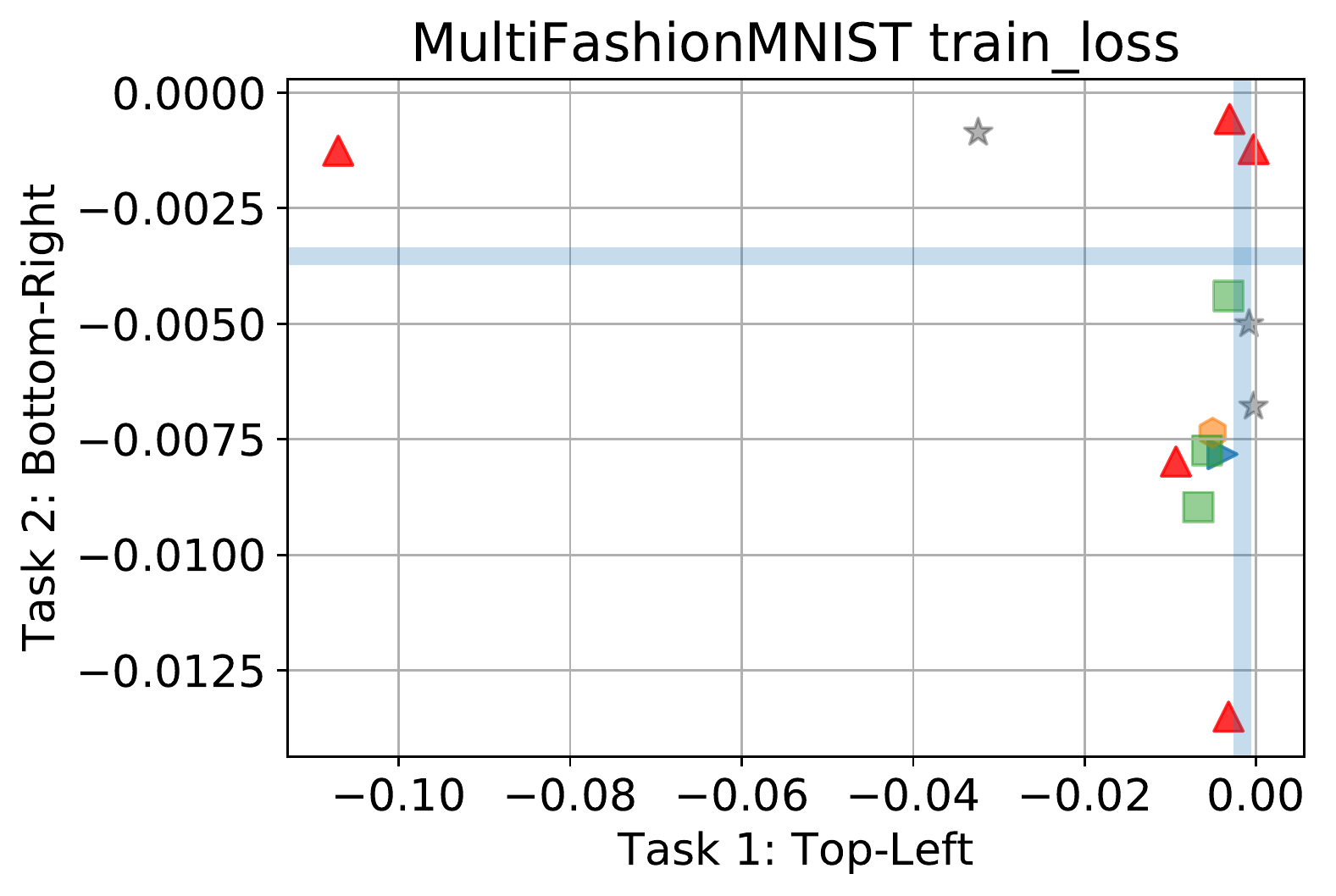}}
\subfloat[MultiFashion: Train Accuracy]{\includegraphics[width = 0.33\textwidth]{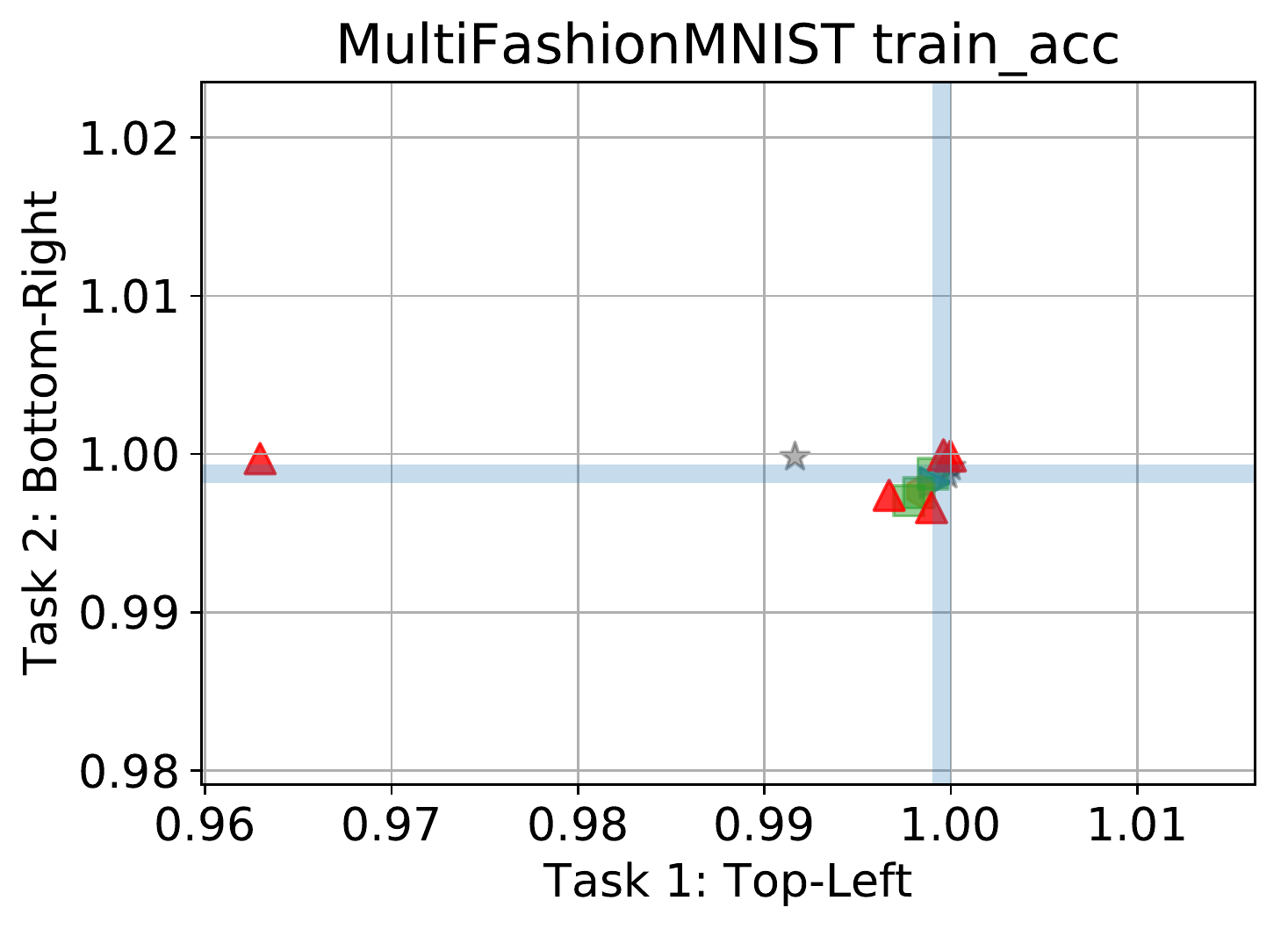}}
\subfloat[MultiFashion: Test Accuracy]{\includegraphics[width = 0.33\textwidth]{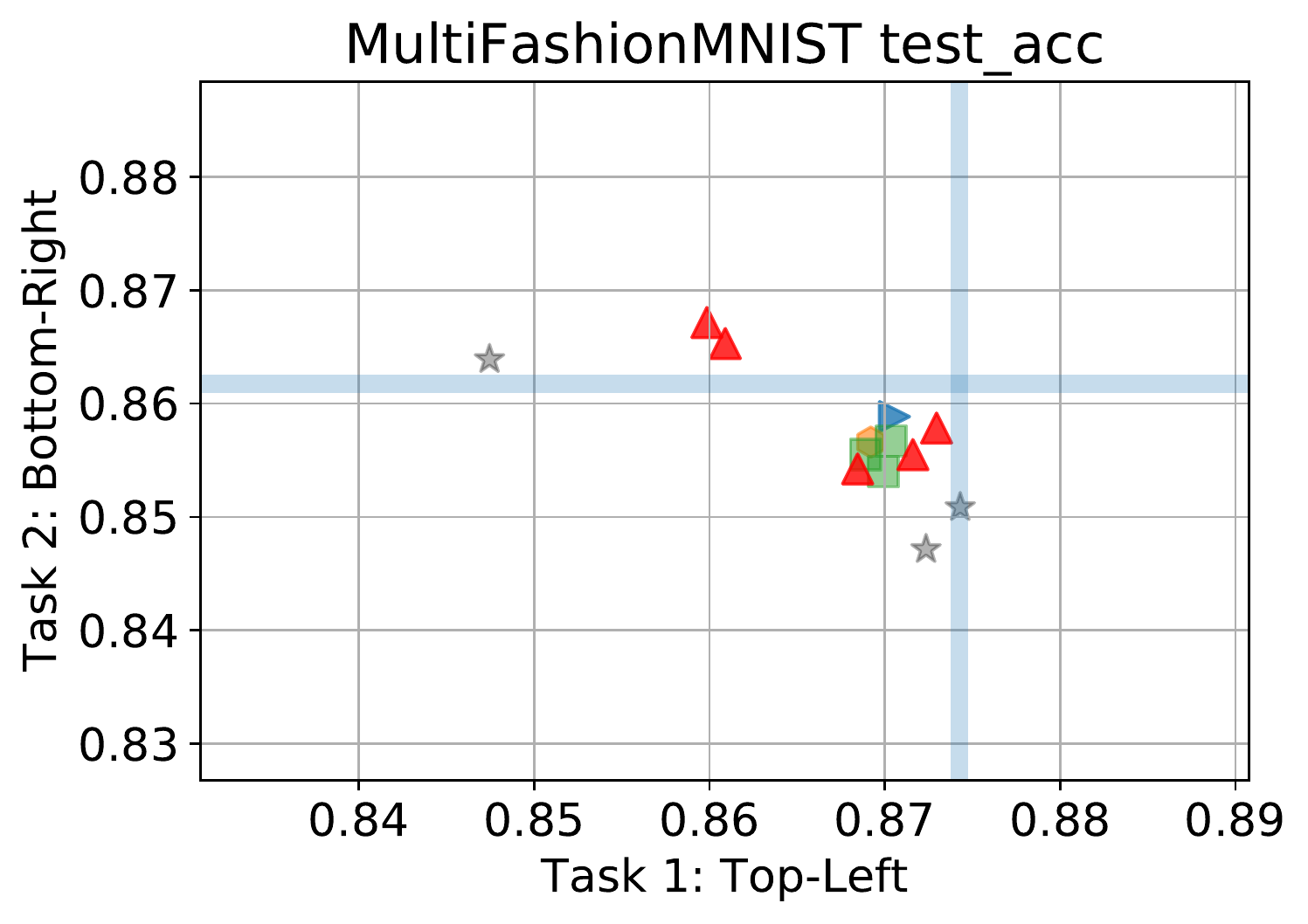}} \\
\subfloat[Fashion\&MNIST: Train Loss]{\includegraphics[width = 0.33\textwidth]{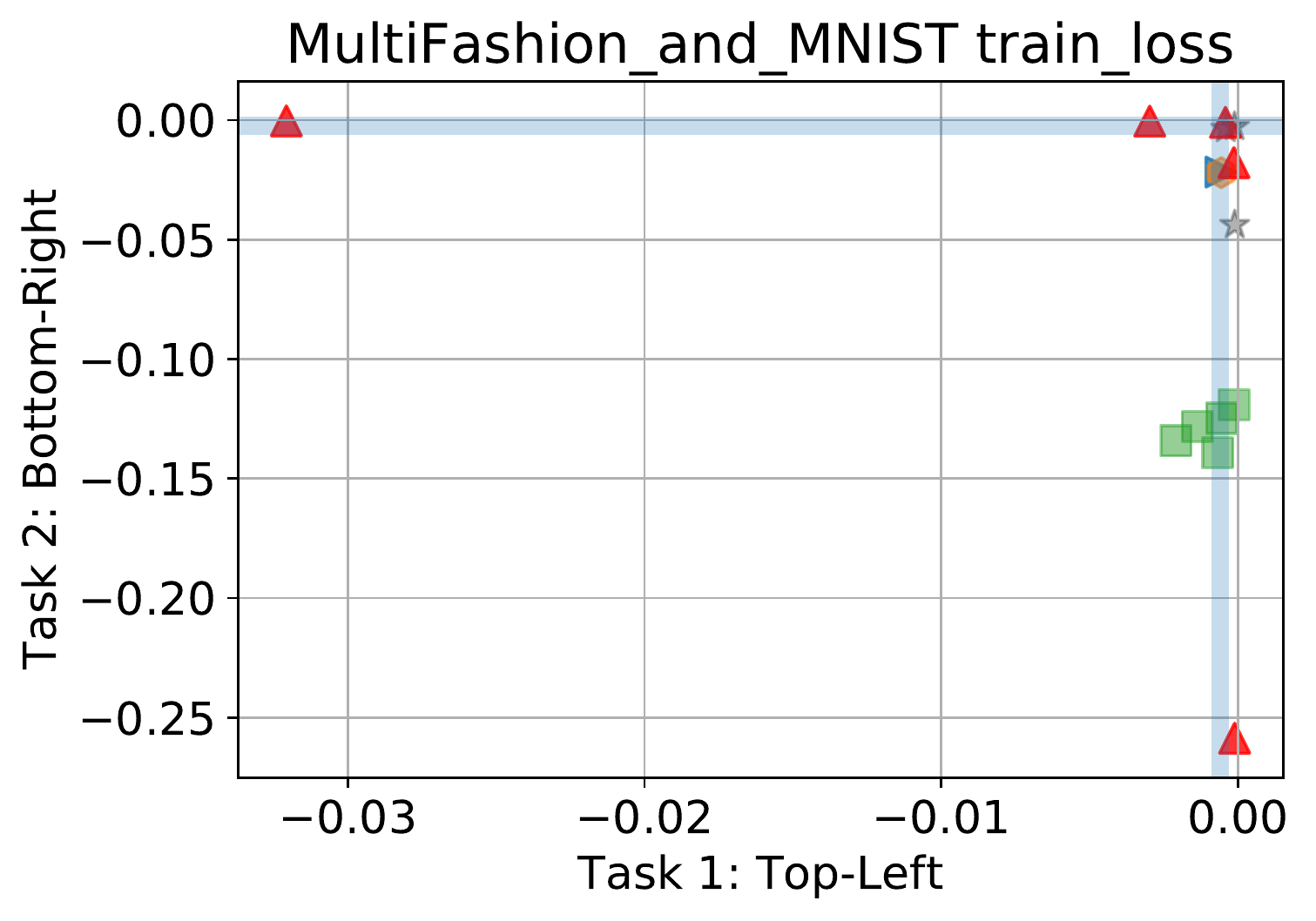}}
\subfloat[Fashion\&MNIST: Train Acc]{\includegraphics[width = 0.33\textwidth]{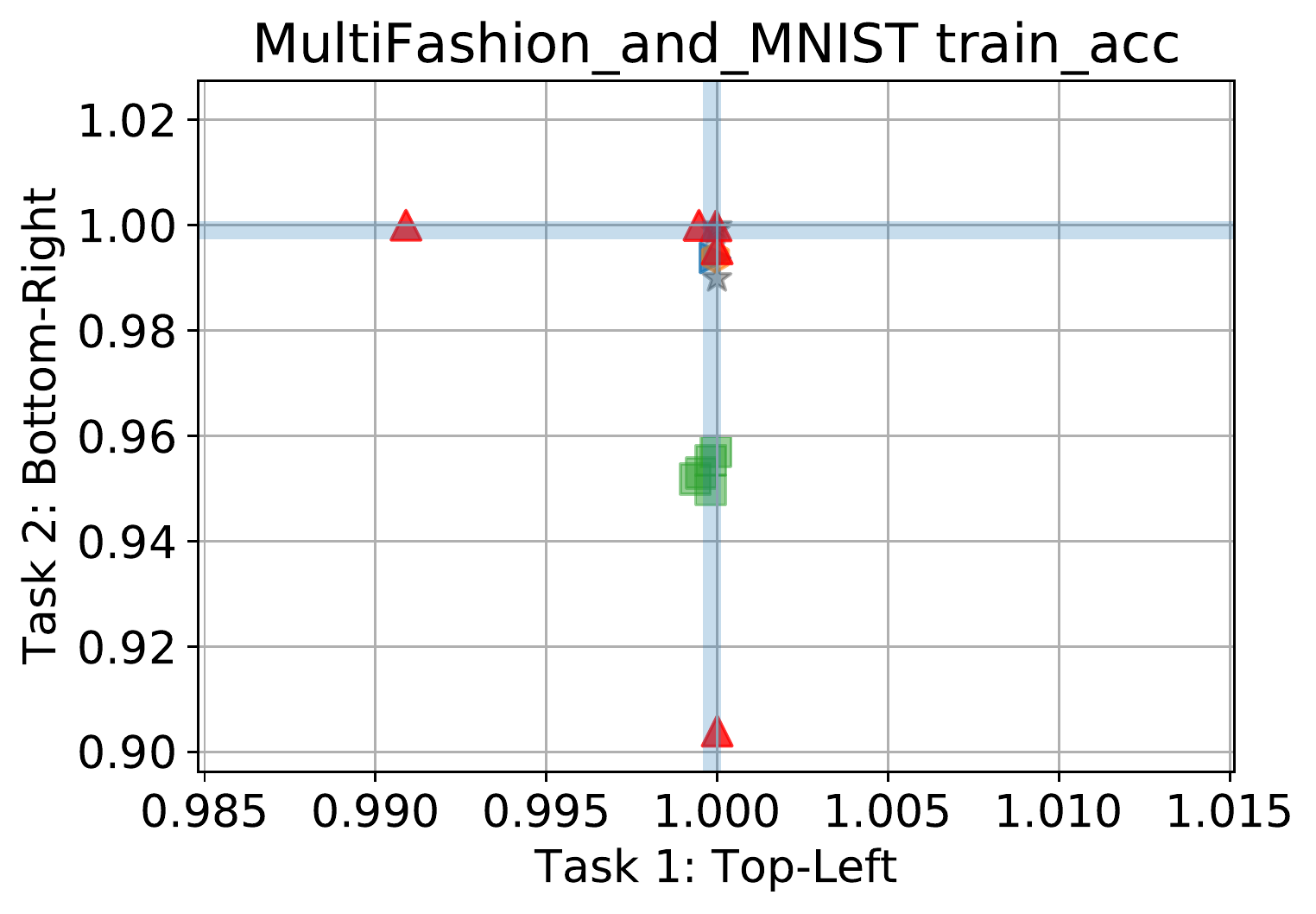}}
\subfloat[Fashion\&MNIST: Test Acc]{\includegraphics[width = 0.33\textwidth]{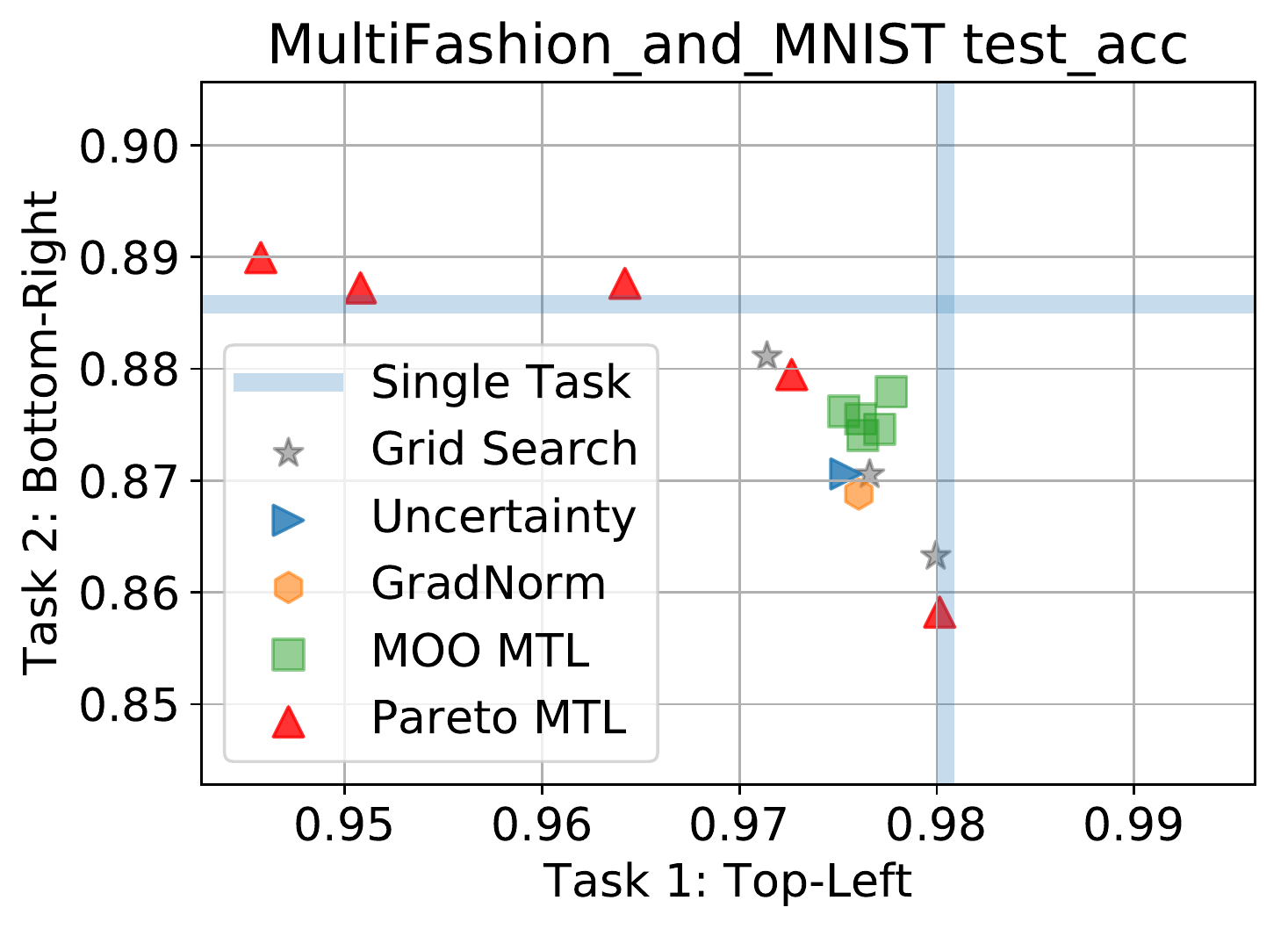}} \\
\caption{The training loss/training accuracy/test accuracy of Pareto MTL and other algorithms with ResNet18 on the MultiMNIST, MultiFashionMNIST and Multi-(Fashion+MNIST) datasets. The labels legend is on the last figure (bottom-right). Pareto MTL has different patterns on the training loss and the test accuracy.}

\label{fig:ResNet18}
\end{figure}

Current work shows that the adaptive gradient methods would have inferior performance on some tasks~\cite{anonymous2020which}. In section 4 of this supplementary material, we have discussed tasks with different difficulty levels which could be one possible reason for the inferior performance. In this section, we provide another discussion based on the gap between optimization and generalization.

Pareto MTL is derived from the view of optimization. To be concrete, we reformulate MTL as a multi-objective optimization problem, and then decompose it into multiple constrained multi-objective subproblems. We also propose an efficient algorithm to solve each constrained subproblem, and treat the obtained solutions as the Pareto solutions for the original MTL problem. However, the objective functions we truly optimize is the \textbf{training loss functions} but not the \textbf{training accuracy} or even the \textbf{test accuracy} for the original MLT problem. Therefore, there would a gap between the objective functions we truly optimize and the MTL generalization ability, which the practitioners most care about. Pareto MTL might have different performance on the training loss and the test accuracy.

To show a clear gap between optimization and generalization on purpose, we test Pareto MTL with ResNet-18 model on the three MNIST-like multi-task datasets, namely the MultiMNIST, MultiFashionMNIST, and Multi-(Fashion+MNIST) in the main paper. The ResNet-18 model could be overparameterized for the MNIST-like dataset and it has the ability to remember all training examples (and hence has a very high training accuracy). To show the different behaviors, we train the models on the MultiMNIST dataset \textbf{with early stop} and train the models on MultiFashionMNIST and Multi-(Fashion+MNIST) \textbf{till the end}.

The experimental results are shown in Fig.~\ref{fig:ResNet18}. For the training losses on all three experiments, Pareto MTL can obtain widely distributed solutions with different trade-offs. This result is not surprising since Pareto MTL makes trade-offs and optimizes the loss functions directly. However, there are different performances on the training accuracy and testing accuracy. For the MultiMNIST dataset \textbf{with early stop}, Pareto MTL has well-distributed solutions on training loss and training accuracy, but they are outperformed by the separate single-task baseline. In contrast, Pareto MTL's solutions on test accuracy are not diverse enough, but they outperform the single-task baseline and provide different optimal trade-off. For MultiFashionMNIST and Multi-(Fashion+MNIST) dataset, we train the models \textbf{till the end}. The training losses for all solutions are close to $0$, and the training accuracy for most solutions are closed to $100\%$. For this extreme case, although Pareto MTL can still generate a set of well-representative solutions on the training loss, it has worse performance on the test accuracy. Some solutions generated by Pareto MTL can match the strong performance of separate single-task baseline on the training loss, and other solutions can provide different optimal trade-offs at the same time. However, they are all outperformed by the single-task baseline on the test accuracy. In other words, Pareto MTL is still good at optimization but has a bad generalization in these cases.

Other adaptive weight methods do not have a clear advantage over the single-task baseline and linear scalarization method with fixed weights, although they sometimes can generate good solutions that match Pareto MTL's solution with the balanced trade-off. In the view of multi-objective optimization, the adaptive weight methods (with proper weight adaption strategy) should be better than linear scalarization with fixed weight. The latter can not find any solution on the concave part of a Pareto front as proved in~\cite{boyd2004convex}. The gap between optimization and generalization could be an important research issue when we design MTL algorithms from the view of optimization.

Another issue for Pareto MLT is the local convergence. As pointed out in the main paper, the solution for each MTL constrained subproblem is restricted Pareto optimal. If the objective functions are all convex and the constraints are properly assigned, Pareto MTL should have a set of widely distributed Pareto solutions. However, especially for training deep neural networks, the loss function would be highly non-convex and the interaction between constrains and the optimization landscape would be complicated. In this case, the solutions for Pareto MTL might be trapped by inferior local Pareto optima. How to get rid of the poor local Pareto optima is an important extension for Pareto MTL.

\section{Conclusion Remark}

In this supplementary material, we provide more experimental results and detailed discussions on Pareto MTL. We also point out several limitations of the current Pareto MTL algorithm and propose some potential research directions. Pareto MTL is derived from the view of multi-objective optimization, and it is orthogonal to many existing MTL methods. We hope there would be further developments on Pareto MTL and its applications for different MTL problems.

\end{document}